# Exploring the effects of Lx-norm penalty terms in multivariate curve resolution methods for resolving LC/GC-MS data

Ahmad Mani-Varnosfaderani *[a], Mohammad Javad Masroor [a]

[a] Chemometrics Laboratory, Department of Chemistry, Tarbiat Modares University, PO-Box: 14115-111

## Abstract

There are different problems for resolution of complex LC/GC-MS data, such as the existence of embedded chromatographic peaks, continuum background and overlapping in mass channels for different components. These problems usually cause rotational ambiguity in recovered profiles and bring uncertainties in the final solutions found using the multivariate curve resolution (MCR) methods. Since mass spectra are sparse in nature, the sparsity constraint has been proposed recently as a constraint in MCR methods for analyzing LC/GC-MS data. There are different ways for implementation of the sparsity constraint, and the majority of methods rely on imposing a penalty term based on the $L_0$-, $L_1$- and $L_2$-norms of recovered mass spectra. Ridge regression and least absolute shrinkage and selection operator (Lasso) can be used for implementation of $L_2$- and $L_1$-norm penalties in MCR, respectively. However, the main question is which Lx-norm penalty method is more worthwhile for implementation of the sparsity constraint in MCR methods. In order to address this question, two and three component LC/GC-MS data were simulated and used for the case study in this work. The areas of feasible solutions (AFS) were calculated using the grid search strategy and fminsearch algorithm. Moreover, the magnitude of the $L_0$, $L_1$- and $L_2$-norms of all mass spectra in AFSs were calculated and visualized as contour plots. The results revealed that the gradient of optimization surface for minimization of $L_1$-norm is much more than those seen for minimization of $L_2$-norm. Therefore, minimization of $L_1$-norm would be a more reliable and practical way for confining AFS and reducing rotational ambiguity for these simulated LC/GC-MS data. Calculating Lx-norms in AFS for $0 \leq x \leq 2$ revealed that the gradient of optimization surface increased from x=2 to x values near zero. However, for x=0, the optimization surface was similar to a plateau, which increased the risk of sticking in local minima. Therefore, the results in this work, recommend the use of $L_1$-norm penalty methods like Lasso for implementation of the sparsity constraint in the MCR-ALS algorithm for finding more sparse solutions and reducing the extent of rotational ambiguity.

* Corresponding Author

Ahmad Mani-Varnosfaderani, email: a.mani@modares.ac.ir, Tel: +98-21-82884738

# 1. Introduction

Recent developments in technologies in analytical chemistry has led to emergence of very complicated instruments, such as two-dimensional gas chromatography-mass spectrometry (GC×GC-MS) [1], GC×GC/MSMS [2], matrix assisted laser desorption ionization imaging mass spectrometry (MALDI-IMS) [3], two-dimensional liquid chromatography-mass spectrometry (LC-LC/MS) [4] and electrophoresis coupled with mass spectrometry (CE-MS and CE/MSMS) [5]. For its high power of characterization of unknown compounds, mass spectrometry is always a very good candidate to be coupled with chromatographic methods and finally make a powerful hyphenated method for separation and identification of compounds in very complex matrices. In recent years, these hyphenated methods have found considerable applications in emerging sciences such as metabolomics, transcriptomics and proteomics [6-8]. Hyphenated methods which include MS for identification are always among the best candidates to be used for solving complex analytical problems. Beyond the high separation and identification power of hyphenated methods, there are always some problems with these instruments, which should be addressed in the analytical community. One of the basic problems with these methods is the huge amount of data generated for a single analytical run. For example, for a single MALDI-IMS run for making mass images of a plant cell, about 10 GB of data is generated, which needs considerable curation and cleaning before the final analysis and interpretation. In fact, high data acquisition rates in modern analytical instruments open a new avenue for analytical chemistry to be considered as a sub-discipline in big data realm [9].

Along with the advances in instrumental technology in analytical chemistry for development of new analytical tools, statistical and data handling algorithms have been growing in recent years for processing of complex data. Different algorithms have been developed by chemometricians for processing of data acquired from hyphenated analytical instruments. Methods like parallel factor analysis (PARAFAC) [10], Tucker decomposition [11] and multivariate curve resolution-alternating least square (MCR-ALS) [12] were proposed and evolved for analysis of three- and two-way chromatographic and spectroscopic data. These methods try to solve different problems in analytical signals, such as the baseline drift, co-elution problem, retention time shift, unknown interferences and problems related to different noise structures in data. Among different factor analysis methods, MCR-ALS has been greatly applied in recent years for analysis of complex and large datasets [13-15]. For analysis of LC/GC-MS data using the MCR methods, different constraints such as non-negativity, unimodality, monotonicity, closure and trilinearity have been proposed. These constraints help a lot to confine the area of feasible solutions (AFS) and find solutions which are more compatible with the physical properties of the investigated system. Recently, another constraint named "sparsity" has been proposed to be implemented in MCR-ALS algorithm for analysis of GC-MS data [16]. This constraint defines some degree of "sparseness" in resolved mass spectra, and it has been shown that implementation of this constraint will help to resolve complicated data matrices [16]. Implementation of this constraint relies on the fact that the mass spectra are naturally "sparse" and thinly dispersed rather than being continuous without any interruption. In 2012, Rasmussen and Bro proposed the sparse version of PARAFAC algorithm for decomposition of three-way data in analytical chemistry [17]. They used $L_1$-norm regularization paradigm in ALS algorithm for sparse decomposition of original data matrix. Least absolute shrinkage and selection operator (Lasso) was used in each iteration of the ALS algorithm for implementation of the "sparsity" constraint in PARAFAC. In a special issue of *Journal of Chemometrics*, Rasmusen discussed the implementation of $L_1$-norm penalty in regular chemometrics algorithms, such as principal component analysis (PCA) and partial least square (PLS) [18]. He concluded that sparse-PCA and sparse-PLS can be applied for analysis of complex data and should be considered when sparse solutions are favorable. Recently, Hugelier et al. proposed a new version of the sparsity constraint in the MCR-ALS algorithm, which was based on regularization of $L_0$-norm of resolved mass spectra [19]. To induce sparseness in solutions, they used a penalized least squares regression framework that constrains the number of the non-null coefficients. They concluded that the application of this $L_0$-norm penalty in the MCR-ALS algorithm increased the chance of finding unique profiles. In another paper, de Rooi et. al. revealed that the degree of sparsity in the $L_x$-norm penalty methods for deconvolution of 1-D data increased from $L_2$-norm to $L_0$-norm [20]. They focused on deconvolution of a vector of pulse trains rather than resolution of a data matrix. Xin-Feng et al. introduced implementation of both $L_2$- and $L_1$-penalty terms in the MCR methods in the architecture of elastic net regression (EN) [21]. In the same way, Cook et. al. used the EN-MCR strategy for sparse spectral recovery [22]. They proposed a strategy for inclusion of both $L_1$- and $L_2$-penalty terms for resolution of LC-MS data. They concluded that EN-MCR returns mass spectra which are superior to those obtained by the MCR-ALS algorithm.

As mentioned, different penalty methods can be used for implementation of sparsity constraint in MCR methods. Regardless of the optimization algorithm used in MCR methods (i.e. ALS, target transformation factor analysis, nonlinear optimization and etc.), the question is, "Which Lx-norm penalty is better for decomposition of a data matrix in terms of the accuracy of the recovered solutions?" The present contribution aims to address this question by investigating the effects of $L_x$-norm constraints on the bands of feasible solutions in the MCR methods for resolution of LC/GC-MS data. For this purpose, two- and three-component LC/GC-MS data have were simulated and used in this study. The results revealed that the implementation of the $L_x$-norm constraint in the MCR methods decreased the extent of rotational ambiguity, especially when $0<x\leq 1$.

## 2. Materials and methods

### 2.1. Sparsity constraint in the MCR methods

Multivariate curve resolution is a set of chemometric techniques which aim to resolve data matrices to spectra and concentration profiles of individual chemical components, given a set of physicochemical constraints. Given a data matrix **D**, an MCR method searches for the best bilinear model as follows:

$$\mathbf{D} = \mathbf{CS^T} \qquad (1)$$

where **C** and **S** are constrained matrices of concentrations and spectra. In most MCR methods, this is done by minimization of the sum of squares (ssq) of the elements of the residual matrix (**E**):

$$\mathbf{E} = \mathbf{D} - \widehat{\mathbf{C}}\,\widehat{\mathbf{S}}^{\mathbf{T}} = \mathbf{D} - \widehat{\mathbf{D}} \qquad (2)$$

$$\text{ssq} = \sum_{i=1}^{n}\sum_{j=1}^{m} e_{ij}^2 \qquad (3)$$

where $e_{ij}^2$ is a typical element of the residual matrix and $\widehat{\mathbf{D}}$ is the predicted data matrix. The MCR-ALS algorithm tries to minimize the ssq by iterative estimation of **C** and **S** matrices using the alternative least square (ALS) approach. The algorithm starts with an initial estimate of **C** or **S** matrices and then tries to calculate the counterpart matrix using the original data (**D**) and the least square technique:

$$\hat{S}^T = (\hat{C}^T C)^{-1}(\hat{C}^T D) \qquad (4)$$

$$\hat{C} = (D\hat{S})(\hat{S}^T\hat{S})^{-1} \qquad (5)$$

The ALS algorithm tries to create a constrained flow of information from **D** matrix to **C** and **S** profiles. Finally, the algorithm continues until the values of the ssq do not change in the successive iterations of the ALS optimization (or the differences between the ssq values will be less than a small threshold, ε).

Since mass spectra are sparse in nature, the implementation of the sparseness constraint on the estimated **S** matrix would be a great idea for confining the possible solutions and obtaining less ambiguous results. This constraint can be applied by minimizing the following objective function instead of the ordinary least squares in the ALS optimization for updating the **S** matrix:

$$\min(\sum_{i=1}^{n}\sum_{j=1}^{m} d_{ij} - \sum_{i=1}^{n}\sum_{j=1}^{m} \hat{d}_{ij})^2 + \lambda \sum_{i=1}^{p}\sum_{j=1}^{m} |\hat{S}_{ij}| \qquad (6)$$

where $\hat{d}_{ij}$ is a typical element of the predicted **D** matrix, p is the number of components and λ is a constant term which determines the degree of importance of the implemented penalty on the $L_1$-norm of the regression coefficients ($\hat{S}$ matrix). The objective function shown in equation (6) is a form of penalized regularization known as "Lasso" [23], which was previously applied to the MCR-ALS algorithm for obtaining sparse solutions for decomposition of GC-MS data [16]. In fact, instead of ordinary least squares (sum of squares of the differences between the original and the predicted data) in equation (4), the above-mentioned objective function (i.e. equation (6)) is applied to each iteration of Lasso-MCR-ALS approach for estimating the sparse spectral profiles ($\hat{S}$). Implementation of this constraint prevents increasing the value of the $L_1$-norm of the predicted

**S** matrix in each iteration of the ALS algorithm. This constraint will lead to more sparse solutions than those usually obtained using the regular least square algorithm.

Different alternatives have been proposed as the second term in equation (6) to define an objective function for solving the regularized least square problem. The elastic net [24] and ridge regression [25] methods are among the most applied algorithms which implement $L_1$- and $L_2$-norm penalties on regression coefficients. The following term is applied to these methods instead of the second penalty term in equation (6):

$$\lambda \left( \sum_{i=1}^{p} \sum_{j=1}^{m} \frac{(1-\alpha)}{2} \left( \hat{S}_{ij} \right)^2 + \alpha \left| \hat{S}_{ij} \right| \right) \qquad (7)$$

The elastic net regression solves the least square problem considering the penalty term mentioned in equation (7) when $0 < \alpha < 1$. As α shrinks toward 0, the elastic net approaches ridge regression and for α=1, the elastic net reaches Lasso regression. Therefore, for α=1, the only implemented penalty term is $L_1$-norm penalty and for α=0, $L_2$-norm penalty is only applied to the least square problem. For $0 < \alpha < 1$, a combination of $L_1$- and $L_2$-norm penalties is applied for estimating the regression coefficients (i.e. **S** matrix in MCR-ALS). Thus, it is possible to tune α parameter to choose between $L_1$- and $L_2$-norm penalties in the elastic net regression when implemented in the MCR-ALS algorithm. Optimization of α and λ parameters is important when working with Sparse-MCR algorithms [22]. The present research is not going to suggest an algorithm for optimization of these parameters in Sparse-MCR-ALS algorithms. This work tries to search the whole space of possible solutions in MCR methods using the grid search strategy and illustrate which solutions are more favorable regarding the value of their $L_x$-norms. Due to the rotational ambiguity in the MCR methods, bands of possible solutions are favorable. This paper probes the solutions in the band and calculates the value of their $L_x$-norms. Then, it is possible to see if the solutions with minimum values of $L_x$-norms are true or not. In order to achieve these goals, the grid search strategy was used for estimating the extent of rotational ambiguity in MCR solutions. All possible solutions of bilinear decomposition were generated using grid search strategy approach and the value of $L_x$-norm for different solutions were calculated.

## 2.2. Grid search approach for estimating the extent of rotational ambiguity

Decomposition of a data matrix can be written in an alternative way using the following equation rather than those presented by equation (1):

$$\widehat{D} = U\, T^T T\, V^T \qquad (8)$$

where U and V are orthogonal scores and loadings of the original data matrix obtained by principal component analysis (PCA) and **T** is a rotation matrix. Due to the resemblance seen between equations (1) and (8), the concentration and spectral profiles can be written as follows:

$$\widehat{C} = U\, T^T \qquad (9)$$

$$\widehat{S} = T\, V^T \qquad (10)$$

For a p component system, **T** is a typical p×p matrix. Since $T^{-1}T$ is an identity matrix, there are infinite numbers of **T** matrices which fit well to equation (8) in order to simulate $\widehat{D}$ very close to the original **D** matrix. In fact, rotational ambiguity is a matter of different **T** matrices which generate $\widehat{D}$ with very small values of ssq. Implementation of constraints on the predicted concentration ($\widehat{C}$) and spectral profiles ($\widehat{S}$) in equations (9) and (10) favors only specific **T** matrices which obey the constraints. It is possible to search the elements of **T** matrix to find the best **T** matrices which fit well with the applied constraints. For a two-component system, **T** is a 2×2 matrix with four elements as follows:

$$T = \begin{bmatrix} T_{11} & T_{12} \\ T_{21} & T_{22} \end{bmatrix} \qquad (11)$$

In order to avoid intensity ambiguity, **T** matrix is normalized. There are different ways for normalizing this matrix. One of the most favorite ways is normalizing it based on diagonal elements to yield the following matrix:

$$T = \begin{bmatrix} 1 & t_{12} \\ t_{21} & 1 \end{bmatrix} \qquad (12)$$

It is possible to search for the best pairs of {$t_{12}$, $t_{21}$} which obey the constraints and yield the least values of ssq. It is rather easy to search this 2-dimensional space for the best pairs of $t_{12}$ and $t_{21}$. Solutions of this problem will make two rectangle (or lines) shapes in this space and the geometric term of location for these pairs is known as area of feasible solutions (AFS) in MCR methods. The favorable AFS for two-component systems can be easily found using a systematic grid search on the values of $t_{12}$ and $t_{21}$ to probe the whole space comprehensively [26].

For a three-component system, the normalized **T** matrix can be written as follows:

$$T = \begin{bmatrix} 1 & t_{12} & t_{13} \\ 1 & t_{22} & t_{23} \\ 1 & t_{32} & t_{33} \end{bmatrix} \qquad (13)$$

Comprehensive optimization of all six elements in the above matrix is a cumbersome task and several alternative approaches have been proposed. A. Golshan et al. proposed a systematic approach to solve the mentioned problem for three-component systems [27]. They systematically searched for the best pairs of {$t_{12}$, $t_{13}$} while optimizing the remaining four elements using a standard simplex algorithm (fminsearch in Matlab). In the present contribution, the same approach was used for optimizing **T** matrix for finding AFS for three-component systems. It is worth mentioning that the non-negativity constraint has been applied for computing AFS using the mentioned approaches for two- and three-component LC/GC-MS systems in this work. In order to avoid intensity ambiguity, all mass spectra in this work were normalized between 0 and 1.

## 2.3. Exploring the effects of L$_x$-norms constraints on the ranges of feasible solutions

As mentioned in the previous section, it is possible to find optimized **T** matrices which obey the constraints and will lead to small values of ssq, using the grid search strategy and fminsearch algorithm. These **T** matrices can be used for defining AFS by plotting pairs of {$t_{12}$, $t_{21}$} or {$t_{12}$, $t_{13}$} against the log(ssq). Each pair of {$t_{12}$, $t_{21}$} or {$t_{12}$, $t_{13}$} defines specific concentrations and spectral profiles. For a 2-dimensional space made by pairs of {$t_{12}$, $t_{21}$} or {$t_{12}$, $t_{13}$}, it is possible to calculate the $L_x$-norm of the spectral profiles for each point and then plot {$t_{12}$, $t_{21}$, $L_x$-norm} or {$t_{12}$, $t_{13}$, $L_x$-norm} as 3-dimensional plots. It is also possible to make a 2D-contour plot for these 3-dimensional plots. In other words, the plot of {$t_{12}$, $t_{21}$} or {$t_{12}$, $t_{13}$} against the $L_x$-norm would be plausible for two- and three-component systems, respectively. Using these plots, it would be possible to see if the spectral profiles with the minimum values of the $L_x$-norms are inside the non-negativity band or not. Moreover, it would be possible to see how the values of the $L_x$-norm of the spectral profiles change inside and outside the non-negativity band. This data will help to better understand the MCR methods with Lx-norm constraints and provides a way for inspecting the reliability of the solutions obtained by these methods.

The LC/GC-MS data with two and three numbers of components were simulated for the case studies. The chromatographic profiles were generated by the exponentially modified Gaussian equation. Moreover, the mass spectra for the simulated GC-MS data were generated with normalized intensities between zero and one. All computer programs involved in this study were coded in-house with MATLAB software (version 8.0.3.532). The calculations were implemented on a server computer with 'Windows 8.1 Enterprise' as the operating system, Intel(R) CPU E5-2690 V3 2.6 GHz (2 processors) and 80 GB of RAM memory.

## 3. Results and discussion

### 3.1. Two-component GC/LC-MS data

The concentration profiles and mass spectra for the two-component GC-MS data in this work are illustrated in Figs. 1a and 1b, respectively. This two-component case study is simulated because of the importance of the overlap usually seen due to the background effect in chromatography data. The mass spectra for the two components do not contain mass peaks in common (i.e. there are no overlaps between them). The grid search strategy was used for finding those elements of **T** matrix which obey the non-negativity constraint for this simulated two-component GC-MS data. The calculated error surface for this data and its contour plot are illustrated in Figs. 2a and 2b, respectively. As can be seen in Fig. 2, there are two distinct regions of feasible solutions for decomposition of this data by implementing the non-negativity constraint. More detailed views of the AFSes are depicted in Figs. S1a and S1b, in the supporting material section. Pairs of {$t_{12}$, $t_{21}$} points which make the true solutions are shown by yellow points in Fig. S1. As can be seen in these figures, there are two rectangle-shaped areas with the minimum and the same values of log(ssq), which indicates the presence of rotational ambiguity in the recovered solutions. In fact, all pairs of {$t_{12}$, $t_{21}$} inside these rectangles are typical spectral/concentration profiles which obey the non-negativity constraint. All of these solutions are mathematically equal and return the same value of ssq. The calculated concentration profiles and normalized mass spectra annotated with all pairs of {$t_{12}$, $t_{21}$} values inside the AFS are illustrated in Fig. S2 in the supplementary material section. Inspection of Fig. S2 reveals that the different shapes of the solutions are favorable and the true solution is one of them. Detailed view of Fig. S2a reveals that many of the recovered concentration profiles for the first component (i.e. background component color coded by red) are different from the true profile, while the concentration profile of the analyte (color coded by blue) is truly recovered. The same issue is also seen for the mass spectra shown in Fig. S2b. As can be seen in this figure, the mass spectra of the background component which is color coded by red is more or less truly recovered, while a large number of artifacts can be seen in different mass channels for the second component (i.e. the analyte color coded by blue). These "non-true" artifacts in the mass spectrum of the analyte component can cause problems when this recovered mass spectrum is supposed to be used for library search in different LC/GC-MS projects. Implementation of other constraints may help to reduce these observed artifacts. The results of this simulation revealed that, in the case of the existence of the continuous background in the concentration profiles in LC/GC-MS data, incorrect determination of the analyte mass spectrum is a possibility. This may cause a lot of mistakes in identification procedures. Both AFSes illustrated in Fig. S1 are equivalent with the identical values of ssq and this twofoldness has been reported previously by Vosough et al. [26]. Due to this symmetrical relationship, only the first AFS shown in Fig. S1a was considered for further analysis in this section. In order to explore the effects of the $L_x$-norm constraints in the MCR methods, the magnitude of $L_2$, $L_1$ and $L_0$ norms of the spectral profiles annotated with pairs of {$t_{12}$, $t_{21}$} in Fig. S1a were calculated. The magnitudes of $L_2$-, $L_1$- and $L_0$- norms for different pairs of {$t_{12}$, $t_{21}$} for the first component of the recovered mass spectra are illustrated as contour plots in Figs. 3a-3c, respectively. The same maps for the second component are shown in Figs. 3d-3f. The non-negativity bands are shown using white dotted lines in these figures. An inspection of Fig. 3 reveals that the magnitudes of $L_2$- and $L_1$-norms of the recovered mass spectra for the first component dramatically decrease from right to left. Moreover, the magnitudes of $L_2$- and $L_1$-norm of the mass spectra of the second component decreased from top to down in Fig. 3d and 3e. Since the true solution was located on the lower left corner of the AFS (shown by a yellow star), minimization of $L_0$-, $L_1$- and $L_2$-norms would be a good suggestion for confining the AFS to reach the true solutions. Interestingly, the magnitude of the $L_0$-norm suffered from a dramatic fall at the left and bottom sides of the AFS for the first and second components, respectively (please see Figs. 3c and 3f). The $L_0$-norm changed from 26 to 14 for the first component from right to left of the AFS. Similarly, the magnitude of the $L_0$-norm changes from 26 to 14 for the second component from the top to the bottom side of the AFS. Large values of the $L_0$-norm in the whole AFS was due to the appearance of the artifacts in different mass channels for both components. Only the true solutions for both components were free of artifacts and therefore the $L_0$-norms for the true solutions were much less than that of other solutions inside the AFS. In order to consider the two components simultaneously; the magnitude of the summation of the $L_x$-norms for the two components (i.e. $\Sigma L_x = L_x$-norm Comp. 1 + $L_x$-norm Comp. 2) for different pairs of $\{t_{12}, t_{21}\}$ are illustrated in Fig. 4a-c respectively for $L_2$, $L_1$ and $L_0$-norms. The true solution was located at the coordinates of (-1.48, 1.49), shown by a yellow star. A comparison of Figs. 4a and 4b reveals that the variation of $\Sigma L_1$ in the AFS was much more sensible than those seen for the values of $\Sigma L_2$, and therefore the minimization of $\Sigma L_1$ was a more practical way for confining the solutions toward the true profiles rather than the minimization of $\Sigma L_2$. The mass spectra which were more similar to the true spectrum were mainly located on the lower left corner of the AFS and also the mass spectra around this area were annotated to smaller values of $\Sigma L_1$. However, the mass spectra with small values of $\Sigma L_2$ still contained solutions which dramatically differed from the true solution (i.e. they were far from the yellow star in Fig. 4a). The iso-$\Sigma L_x$ lines are shown as dotted gray lines in Fig. 4. The iso-$\Sigma L_2$ lines are the convex lines whose curvature increase from right to left. However, the iso-$\Sigma L_1$ lines are made by different line segments and make different angles with each other. The iso-$\Sigma L_0$ lines are also direct line segments and the angles between them are equal to 90 degrees. In fact, for $1<x\leq2$ the iso-$\Sigma L_x$ lines are convex and for $0<x<1$ the iso-$\Sigma L_x$ lines are concave. This phenomenon was visualized by a movie named "movieS1.mp4", available in the supplementary material section, which animates the variation of the magnitude of the $\Sigma L_x$-norm against the values of $t_{21}$ and $t_{12}$ in mesh plots as a function of x. The axis of mesh plots were scaled to the maximum and minimum of the $L_x$-norms in this movie. From the optimization point of view, two different variations have to be considered in this movie. The first is the gradient of the changes of the $\Sigma L_x$-norms near the true solution (i.e. the coordinates of $t_{12}$=-1.48 and $t_{21}$=1.49) and the second is the variations of the $\Sigma L_x$-norms inside the whole AFS. Different frames of this movie revealed that for x values close to 2, the gradient of the $\Sigma L_x$-norm near the true solution was not too much but it increased as x changed from 2 to 0. For x values less than 0.5, the gradient of the changes of the $\Sigma L_x$-norms near the true solution increased dramatically. It means that the optimization methods could easily find the true solution when 0<x<0.5. However, for x=0, the gradient of the changes of the $\Sigma L x$-norms in the whole AFS was zero. In this special case, which is animated in the final frames of the movie, the norm surface inside the AFS became a plateau. There was no gradient inside the AFS. Therefore, the optimization methods considering the $\Sigma L_0$- norm of spectra as fitness function may stick in local minima and cannot find the best solution on the left corner of the AFS.

Another two-component LC/GC-MS data was simulated with some degrees of overlapping in the mass channels. The same procedure was used in order to investigate the effects of the $L_x$-norms on rotational ambiguity. The concentration and spectral profiles together with the results are given in Appendix A. Generally, the same conclusions were obtained; however, an interesting difference with the previous example was observed. Detailed inspection of Figure A5 reveals that, when some degrees of overlapping exist in the mass channels, the minimum values of $\Sigma L_1$ and $\Sigma L_2$-norms do not coincide in the same position in the AFS. In fact, the minimum $\Sigma L_2$-norm does not represent the true solution, while for the $\Sigma L_x$ norms with 0<x≤1, the minimum of $\Sigma L_x$ and the true solution are in the same place in the AFS. This is illustrated using a movie named "movieA.mp4" in the supplementary material section. This observation implies that the minimization of the $\Sigma L_2$-norm would not be a good suggestion for the implementation of the sparsity constraint in the MCR methods for resolution of LC/GC-MS data, especially when some degrees of overlapping exists in mass channels.

Similar to the results obtained in this work, Hugelier et al. [19] mentioned that solving $L_0$-norm optimization in MCR is a complex, non-convex and non-differentiable optimization task, and can only be solved by using an approximation of the $L_0$-norm. They used an iterative process starting from the unweighted ridge regression until convergence of the solution for finding the spectrum with minimum $L_0$-norm value [19]. The approximation algorithm proposed by Hugelier et al. works for different chemical components, separately. In fact, they developed a deconvolution method and imposed it inside the ALS algorithm for each component. They did not consider this minimization for $L_0$-norm of several chemical components, simultaneously. This issue opens an avenue for development of new $L_0$-norm optimization methods which works for multivariate data matrices rather than a vector of signal intensities. The results in this work, reveals that the solutions of minimum $L_1$-norm coincide with the solutions of $L_0$-norm, but the $L_1$-norm optimization problem is differentiable and can be solved using existing algorithms such as Lasso even in multivariate case. Minimization of $L_1$-norm of mass spectra can be achieved simultaneously for several components, together.

### 3.2. Three component GC/LC-MS data

The concentration profiles and mass spectra for the simulated three-component GC/LC-Ms data in this work are illustrated in Fig. S3a and S3b in the supplementary material section. Using the strategy discussed in section 2.2 and the optimization of **T** matrix in equation (13), an error surface which is a mesh plot of the pairs of $\{t_{12}, t_{13}\}$ against the log(ssq) was obtained and illustrated in Fig. 5a. Inspection of this figure reveals that there are three distinct regions in $t_{12}$-$t_{13}$ coordinates which fulfill the data and non-negativity constraint. The detailed views of these three AFSes are shown in Figs. 5b-d. These three regions were named AFS-I, AFS-II and AFS-III, respectively (please see Fig. 5). Each pair of $\{t_{12} t_{13}\}$ inside the non-negativity band is a solution. Fig. S4 shows all the recovered concentration and spectral profiles for the different pairs of $\{t_{12} t_{13}\}$ in the three AFSes illustrated in Fig. 5. Similar to those seen for the two-component system, different sets of the concentration profiles and mass spectra fitted the data with the non-negativity constraint and there were bands of non-negative profiles rather than a unique solution. The AFS-I, II and III were responsible for all the solutions seen in Fig. S4 for the first, second and third components, respectively. A detailed analysis of Fig. S4 reveals that the recovered mass spectra for the first and second components are mixed-up with the mass spectrum of the background component (color coded by red). The presence of this uncertainty is due to the existence of rotational ambiguity in this three-component data matrix. As discussed in the previous section, minimization of $L_x$-norm of spectral profiles can be considered as a systematic way for reduction of rotational ambiguity in this decomposition problem. In order to explore the selection of the best $L_x$-norm as the objective function for this minimization problem, the $\Sigma L_x$-norms of the mass spectra for three AFS regions were calculated. The $\Sigma L_2$-, $\Sigma L_1$- and $\Sigma L_0$-norms of the mass spectra in AFS-I as a function of $t_{12}$ and $t_{13}$ are illustrated as contour plots in Figs. 6a-c, respectively (the high resolution versions of these figures are given in the supplementary material section). The yellow star point represents the true solution for the first component in AFS-I. Inspection of Figs. 6a and 6b reveals that both $\Sigma L_2$- and $\Sigma L_1$- norms are minimized near the true solution in the coordinates of the $t_{12}$-$t_{13}$ space. The iso-$\Sigma L_2$ lines in Fig. 6a are concentric ovals and the true solution is at the center of all iso-$\Sigma L_2$ lines. A detailed view of Fig. 6b reveals that the iso-$\Sigma L_1$ lines in this figure are concentric lozenge shapes with the true solution at the center point. Comparison of Figs. 6a and 6b reveals that the gradient of the $\Sigma L_1$- norms near the true solution in the $t_{12}$-$t_{13}$ space is much more than those of the $\Sigma L_2$- norms. It suggests that the optimization of the $\Sigma L_1$-norm is a better way for finding the true solution in AFS-I rather than the minimization of the $\Sigma L_2$-norm. An inspection of the $\Sigma L_0$-norm for AFS-I in Fig. 6c reveals that, for the whole $t_{12}$-$t_{13}$ space, the magnitude of the $\Sigma L_0$-norm is constant and equal to 123. It shows that the optimization of the $\Sigma L_0$-norm is not a good suggestion for finding the true solution or confining the nonnegativity band seen for AFS-I. The magnitudes of the $\Sigma L_2$-, $\Sigma L_1$- and $\Sigma L_0$-norms of the mass spectra in AFS-II in the coordinate $t_{12}$-$t_{13}$ space are illustrated in Figs. 6d-e, respectively. Similar to those seen for the two-component example and AFS-I, the iso-$\Sigma L_2$ lines are concentric ellipsoids and the true solution (shown by a yellow star) is located at the center. Similarly, the iso-$\Sigma L_1$ lines are lozenge shaped. A comparison of Figs. 6d and 6e reveals that the gradient of the $\Sigma L_1$-norms near the true solution in the $t_{12}$-$t_{13}$ coordinate is much more than those seen for the $\Sigma L_2$-norms in AFS-II. Therefore, the optimization of the $\Sigma L_1$-norm would be a good suggestion for confining AFS-II and reduction of rotational ambiguity. The values of the $\Sigma L_2$-, $\Sigma L_1$- and $\Sigma L_0$-norms of the mass spectra in AFS-III are color-coded in Figs. 6g-h as contour plots, respectively. As illustrated in these figures, the magnitudes of the $\Sigma L_2$-, $\Sigma L_1$- and $\Sigma L_0$-norms in the $t_{12}$-$t_{13}$ coordinates of AFS-III are constant and there is no gradient inside AFS-III. This reveals that the optimization of the $L_x$-norms in this AFS cannot help for confining the solutions and obtaining more accurate results. AFS-III is the collection of the solutions for the third component (the background component color coded by red) and the size of AFS-III is much smaller than that of AFS-I and AFS-II. In fact, for this three-component case study in this work, the minimization of the $\Sigma L_2$-, $\Sigma L_1$-norms will help to obtain the unique and true solution for the first and second components. General view of Fig. 6 reveals that the minimization of the $\Sigma L_0$-norm cannot help for finding true solution and confining the AFSes. As can be seen in Figs. 6c, 6f and 6i, the values of the $\Sigma L_0$-norms in all sections of the $t_{12}$-$t_{13}$ coordinates are constant and equal to 123. Therefore, for the simulated three-component GC/LC-MS data in this work, the $\Sigma L_0$-norm will not help to reduce rotational ambiguity because the optimization surface is a complete plateau.

It is worth mentioning that the artifacts seen in Figs. 6d, e, g and h are due to the immature optimization of **T** matrix using the fminsearch approach. For a better optimization of **T** matrix we increased the resolution of the grid search approach up to the limits of the RAM memory (80 GB). The best figures are those shown in Fig. 6. Images with higher resolution required more than 80GB of RAM memory.

In order to explore overlapping in mass channels, other three-component LC/GCMS data was simulated with three Gaussian overlapping chromatograms and some degrees of overlapping in mass channels. The same procedure was used in order to investigate the effects of Lx-norms on rotational ambiguity. The results are given in Appendix B. The contour plots for the $\Sigma L_2$-, $\Sigma L_1$- and $\Sigma L_0$- norms in the $t_{12}$-$t_{13}$ space for AFSI, AFSII and AFSIII are illustrated in Fig. B4. Detailed view of this figure reveals that the gradient of the changes for the $\Sigma L_1$-norm is more than those seen for the $\Sigma L_2$-norm for AFSI, AFSII and AFSII. This suggests that optimization of $\Sigma L_1$-norm is a more practical way for finding the true solution. Moreover, for AFSI and AFSIII, the minimum values of the $\Sigma L_1$- and $\Sigma L_2$-norms do not coincide in the same position. In contrast to the $\Sigma L_2$-norm, the minimum values of the $\Sigma L_1$-norm are in the same position as the true solution (the yellow star points). This again implies that minimization of $\Sigma L_1$-norm of mass spectra is more reliable for finding true and sparse solutions than those of $\Sigma L_2$-norms. The results in Appendices A and B imply that for the LC/GC-MS data with overlapping in the mass channels, the minimum values of the $\Sigma L_1$- and $\Sigma L_2$-norms of the mass spectra are not in the same position in AFSes. The same conclusion about the $\Sigma L_0$-norm was derived in this section compared to the results given for the other two- and three-component examples.

## 4. Conclusion

As discussed in section 2.1., the sparsity constraint can be implemented in the MCR methods using different algorithms and penalty functions. Hugelier et al. [19] used a least squares regression framework with $L_0$-norm penalty for implementation of the sparsity constraint in MCR solutions for resolution of imaging mass spectroscopy data. Pomareda et al. [28] used the $L_1$-norm constraint in resolution of ion mobility spectra by implementation of Lasso regression techniques in MCR-ALS algorithm. Moreover, Hugelier developed an $L_1$-constrainted MCR-ALS algorithm for deconvolution of high-density super-resolution images [29]. This method implements Lasso regression algorithm instead of ordinary least square in MCR-ALS for recovering more sparse profiles in images. Tikhonov regularization [30] and ridge regression [25] both implement the $L_2$-norm constraint and can be used in MCR-ALS algorithm for finding sparse solutions. Generally, there are different algorithms for solving the bilinear decomposition problem mentioned in Eq. (1) by implementation of $L_0$- $L_1$- and $L_2$-norm penalties. The results obtained in this work suggest that optimization of $\Sigma L_1$-norm of mass spectra would be a more practical way for implementation of the sparsity constraint and finding the true solution when analyzing complex LC/GC-MS data. When optimizing $\Sigma L_2$-norm using Eq. (7) for $\alpha=0$, the gradient of the $L_2$-norm in the optimization surface is gradual and not too sharp near the true solution. However, optimization of $L_x$-norm would be beneficial when $0 < x \leq 1$. Moreover, the results for two- and three-component LC/GC-MS data with overlapping in the mass channels revealed that the minimum values of the $\Sigma L_1$- and $\Sigma L_2$-norm did not coincide in the same position in AFSes. In these examples, minimization of the $\Sigma L_1$-norm was more reliable than the $\Sigma L_2$-norm for finding the true solution. The calculations in this work revealed that optimization methods implementing the $L_0$-norm constraint may not be very useful for resolution of LC/GC-MS data and there are some risks for sticking in local minima due to the complete flattened optimization surface. Since the Lasso algorithm is a standard and practical way for implementation of the $L_1$-norm constraint in the least square problem, implementation of this algorithm in each iteration of MCR-ALS is recommended for resolution of complex LC/GC-MS data.

## Conflict of interest

The authors declare no conflict of interest.

## Acknowledgments

The authors would like to thank Tarbiat Modares University for the financial support of this project.

**Figure Captions**

**Figure 1.** (a) The concentration profiles and (b) mass spectra for the simulated two component LC/GC-MS data.

**Figure 2**. (a)The error surface and (b) contour map of log (ssq) against $t_{12}$ and $t_{21}$ obtained using grid search strategy for implementation of "non-negativity" constraint for the simulated two component LC/GC-MS data.

**Figure 3**. (a) The contour plots of the (a) $L_2$-norm, (b) $L_1$-norm and (c) $L_0$ norms of the mass spectra in $t_{12}$-$t_{21}$ space for the first component of the simulated LC/GC-MS data. The counter plots of the (d) $L_2$-norm, (e) $L_1$-norm and (f) $L_0$-norms of the mass spectra in $t_{12}$-$t_{21}$ space for the second component of the simulated LC/GC-MS data.

**Figure 4**. The contour plots for (a) $\Sigma L_2$-, (b) $\Sigma L_1$- and (c) $\Sigma L_0$- norms in $t_{12}$-$t_{11}$ space for AFSI for the simulated two component LC/GC-MS data.

**Figure 5**. (a) The error surfaces obtained using grid search strategy and fminsearch approach for implementation of "non-negativity" constraint for the simulated three component LC/GC-MS data. (b) Contour plot of AFSI, (c) Contour plot of AFSII and (d) Contour plot of AFSIII

**Figure 6**. Contour plots for (a) ΣL$_2$-, (b) ΣL$_1$- and (c) ΣL$_0$- norms in t$_{12}$-t$_{13}$ space for AFSI, (d) ΣL$_2$-, (e) ΣL$_1$- and (f) ΣL$_0$- norms in t$_{12}$-t$_{13}$ space for AFSII and (g) ΣL$_2$-, (h) ΣL$_1$- and (i) ΣL$_0$- norms in t$_{12}$-t$_{13}$ space for AFSIII.

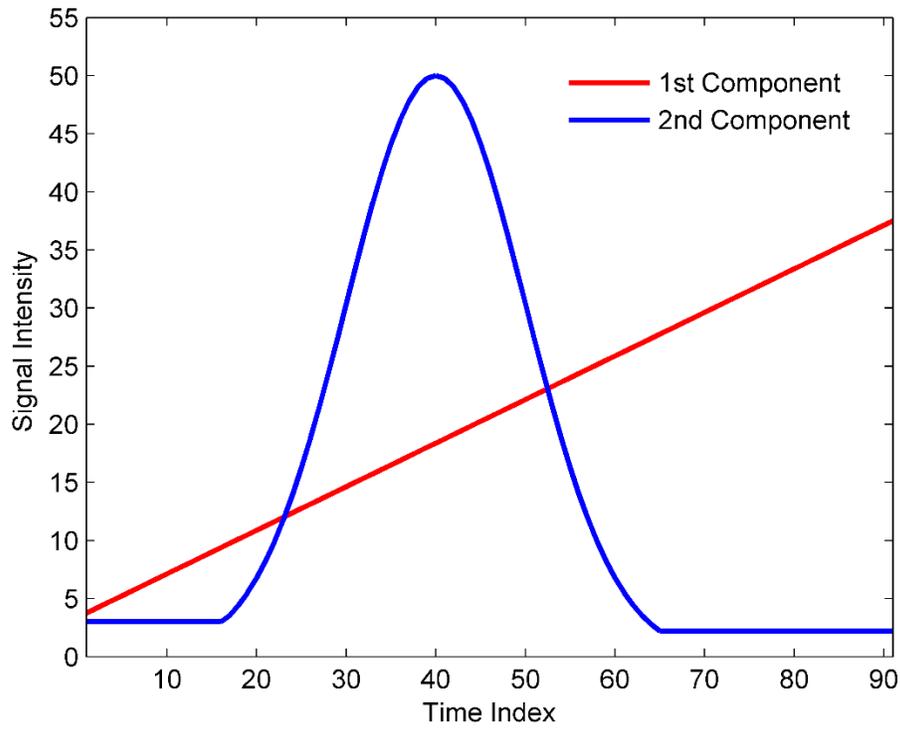

Figure 1a.

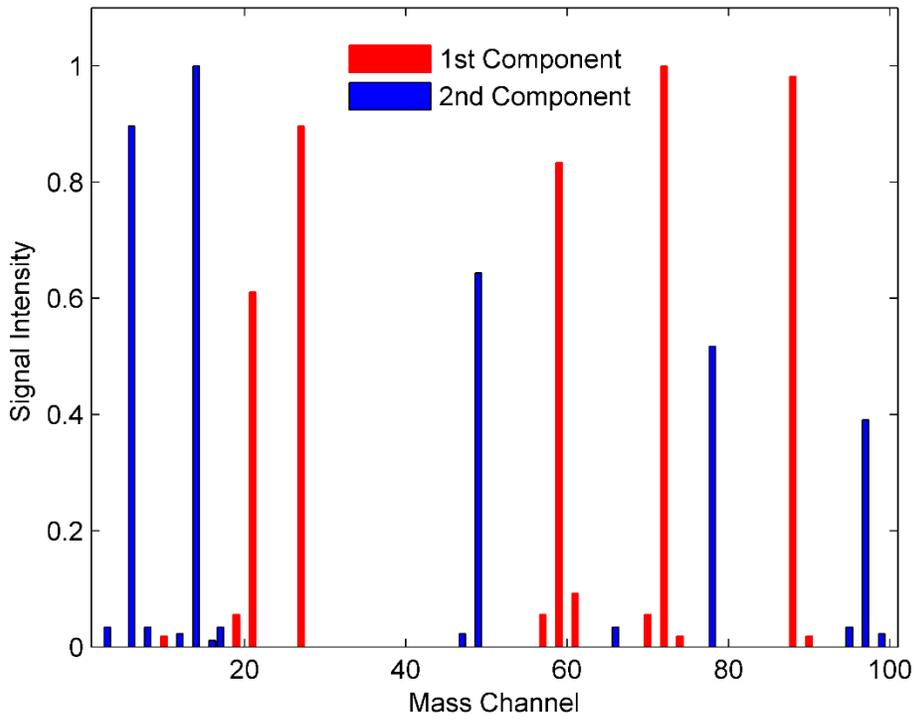

Figure 1b.

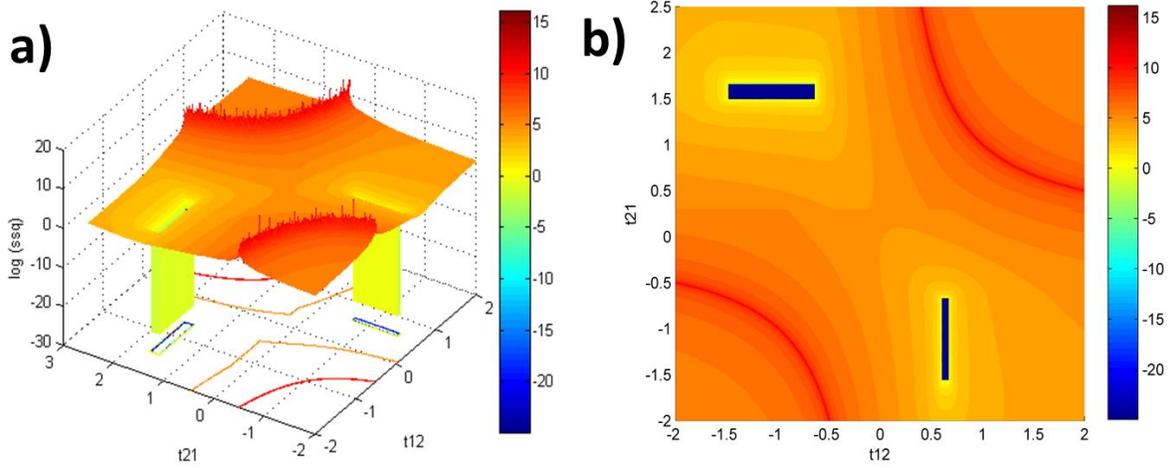

Figure 2.

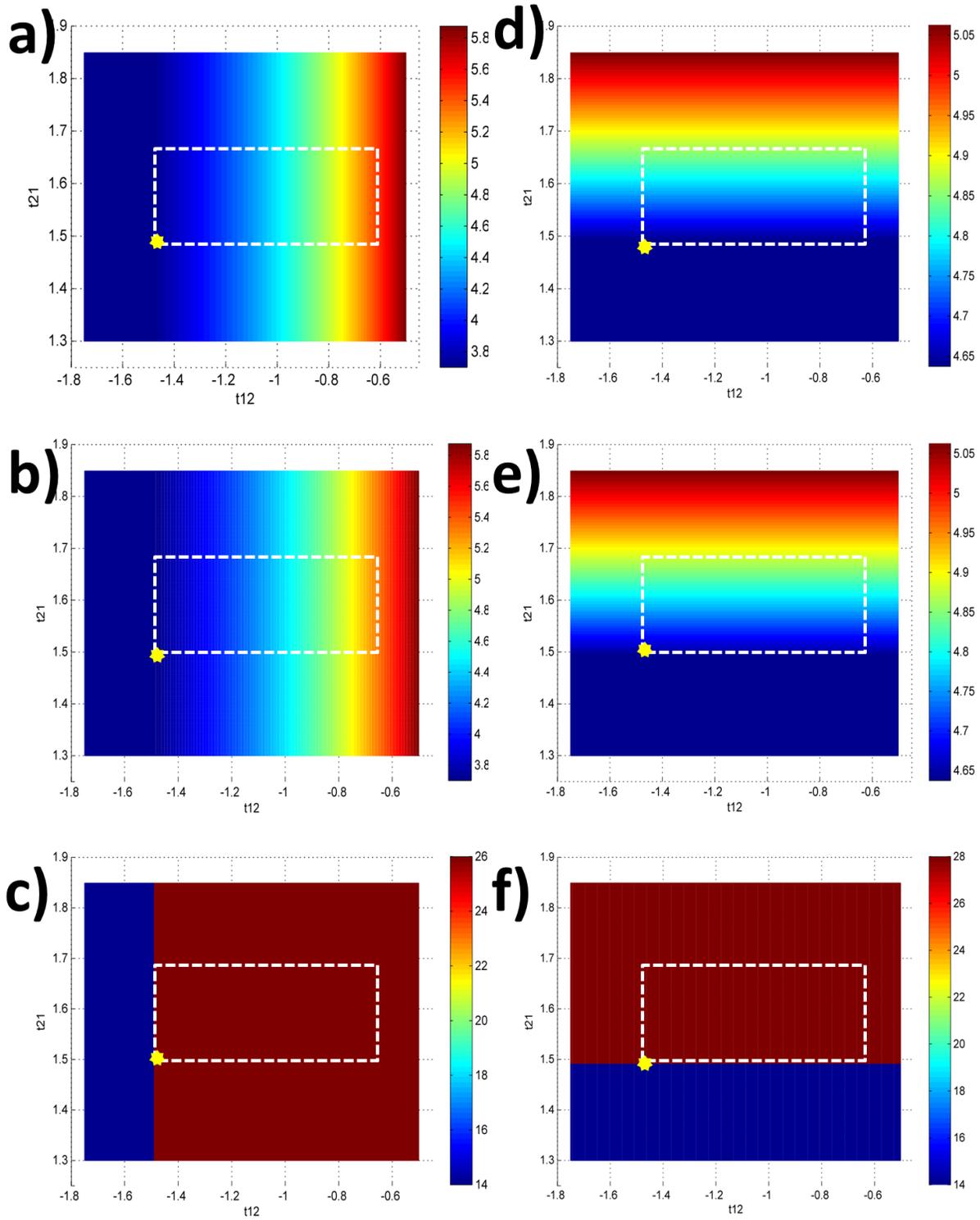

Figure 3.

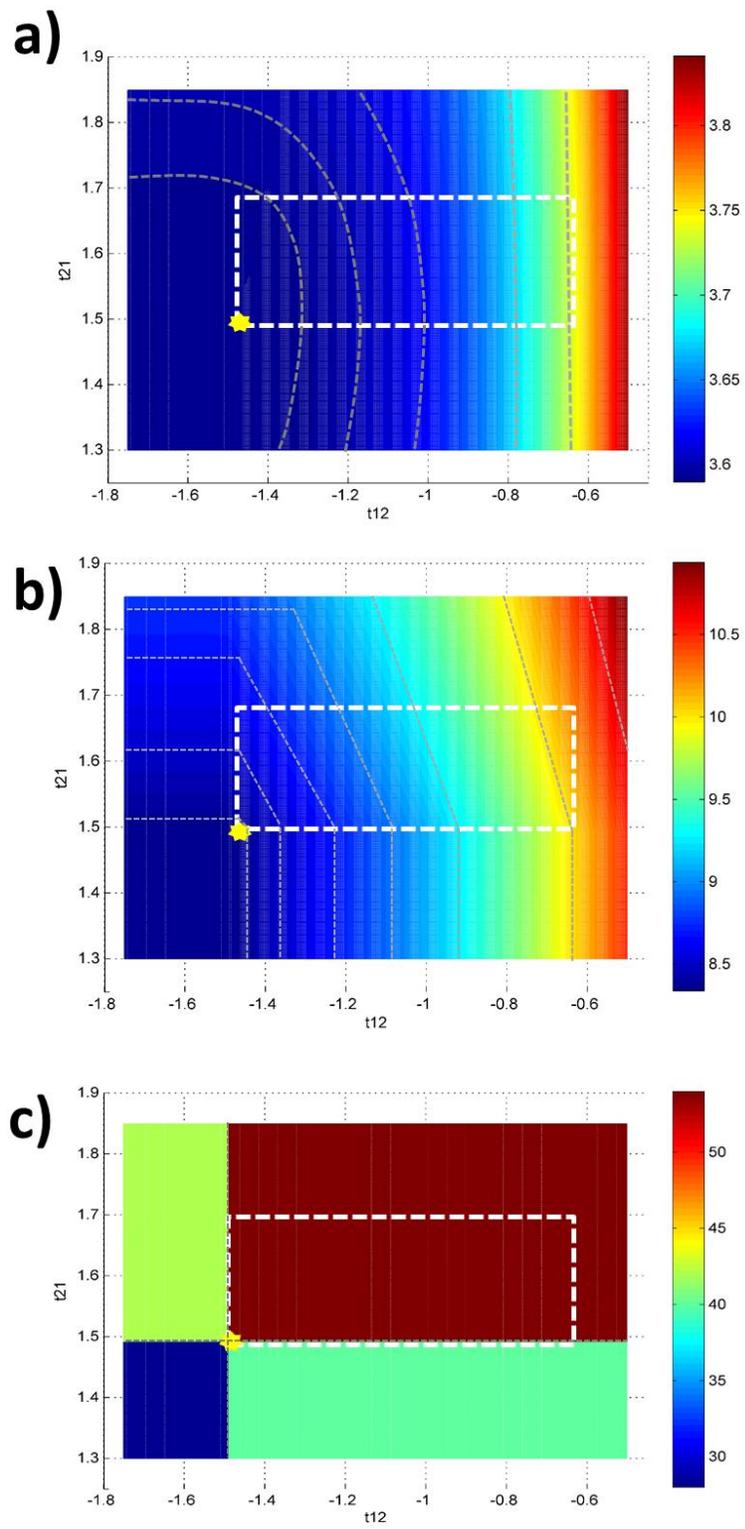

Figure 4.

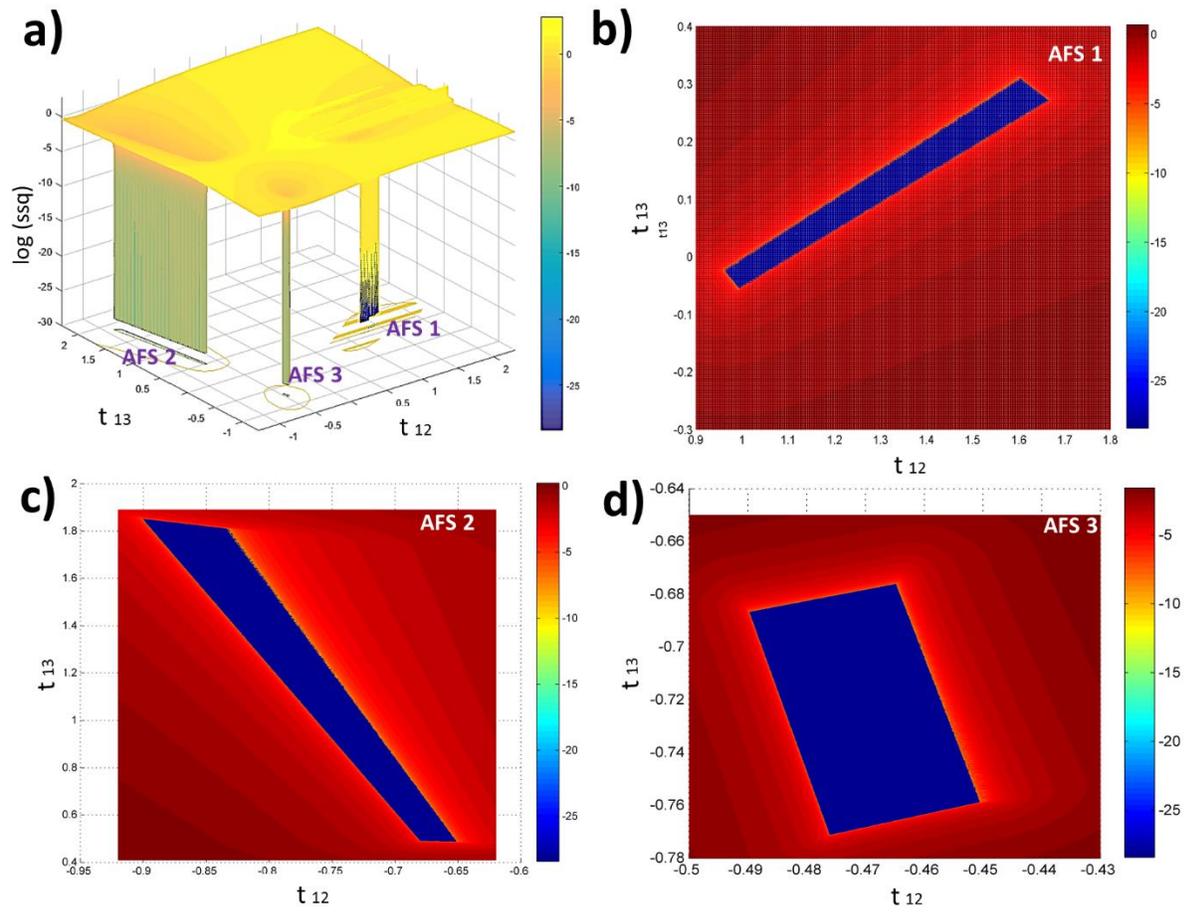

Figure 5.

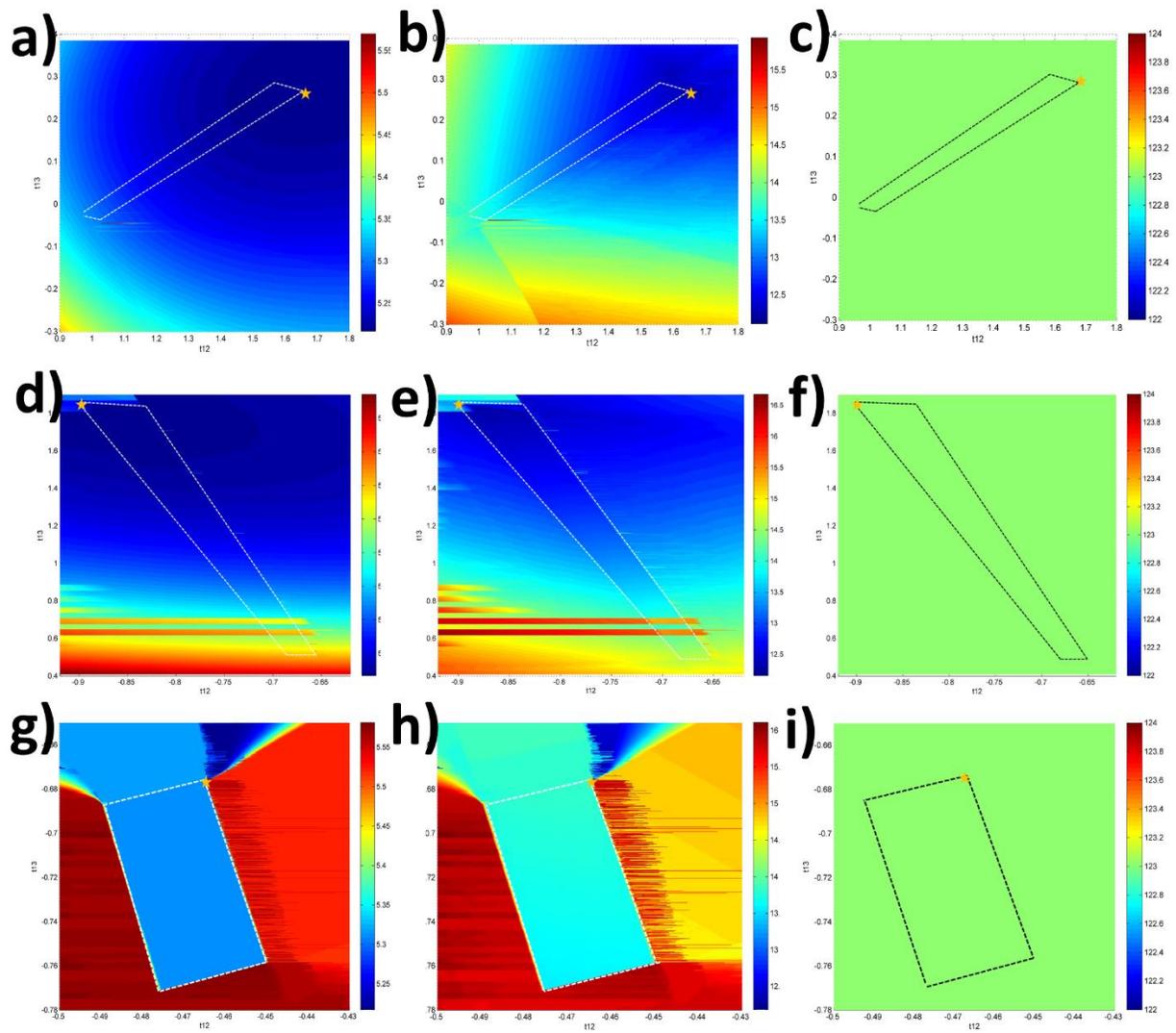

Figure 6.

# Appendix A

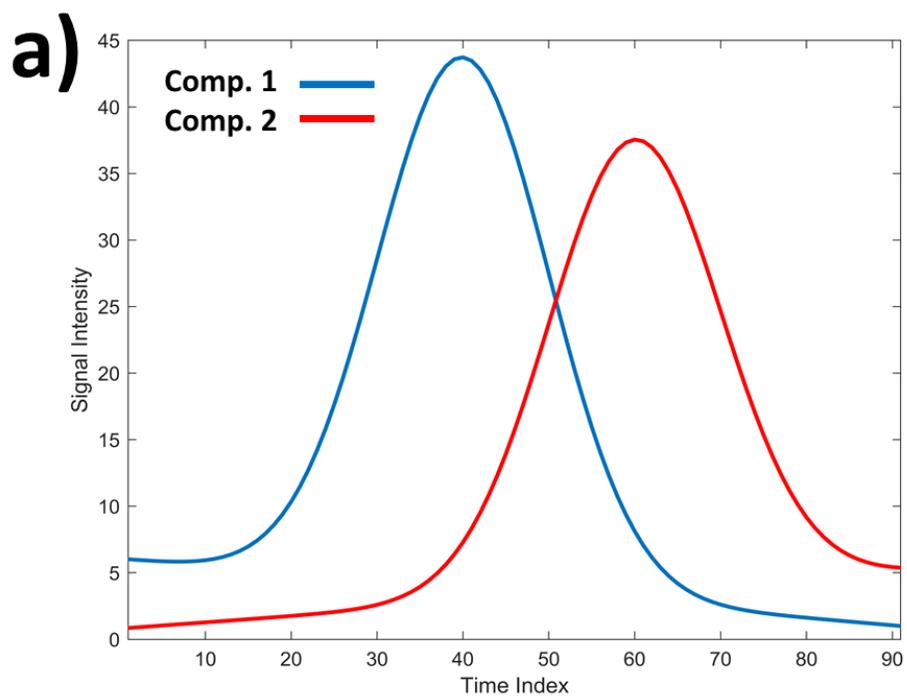

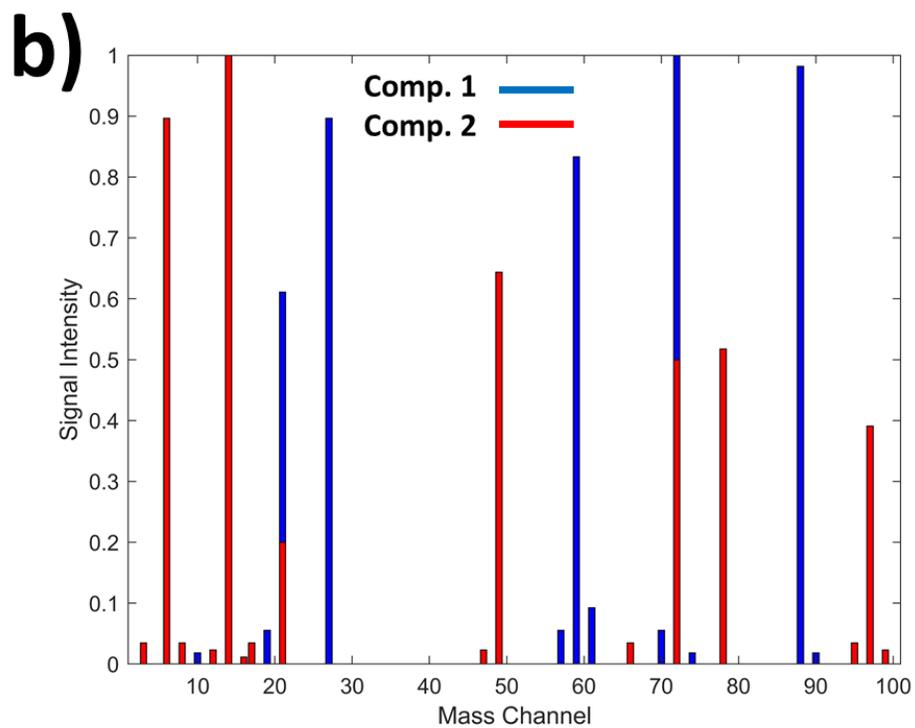

**Figure A 1.** (a) The concentration profiles and (b) mass spectra (with overlapping in mass channels) for the simulated two component LC/GC-MS data in Appendix A.

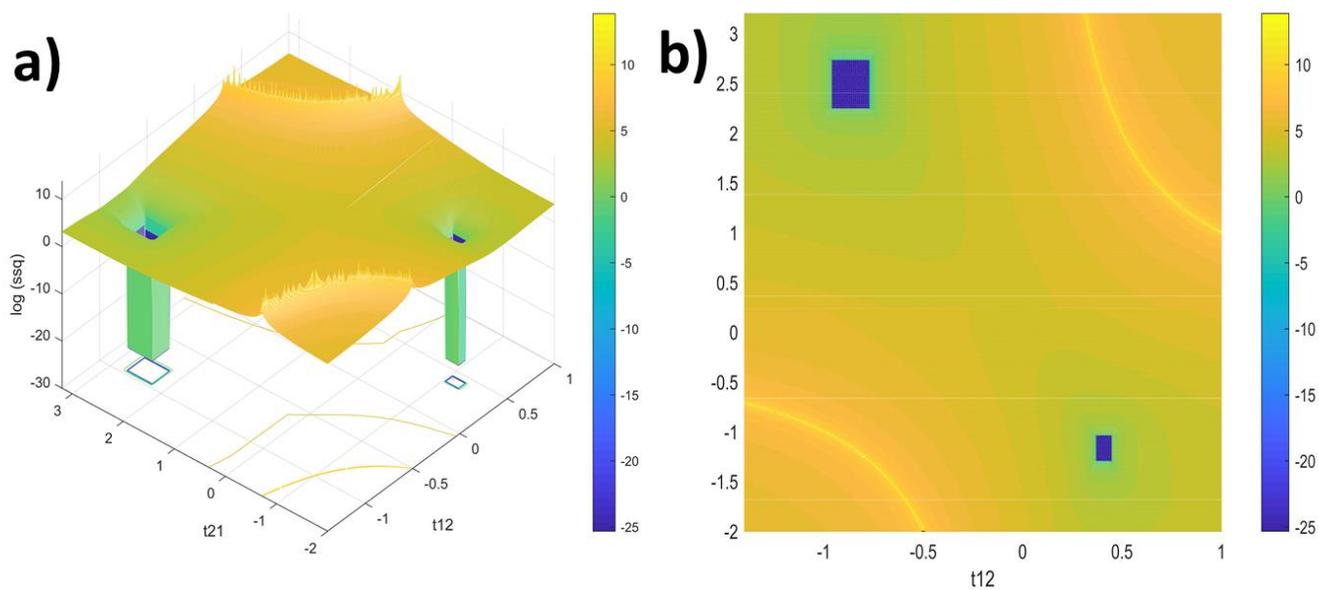

**Figure A2**. (a)The error surface and (b) contour map of log (ssq) against $t_{12}$ and $t_{21}$ obtained using grid search strategy for implementation of "non-negativity" constraint for the simulated two component LC/GC-MS data in Appendix A.

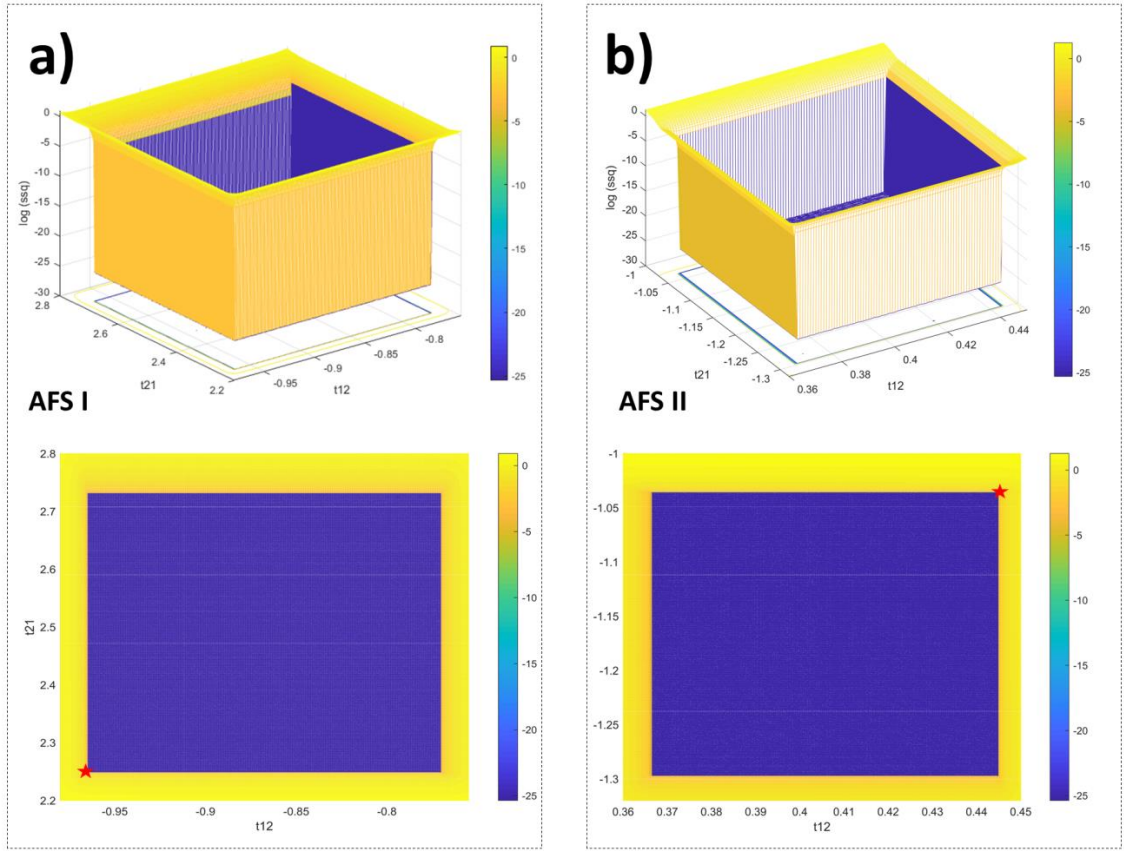

**Figure A3**. The plot of the log (ssq) against $t_{12}$ and $t_{21}$ together with their contour plots for the simulated two component system in Appendix A: (a) AFSI and (b) AFS II. Red star points reveal the true solutions.

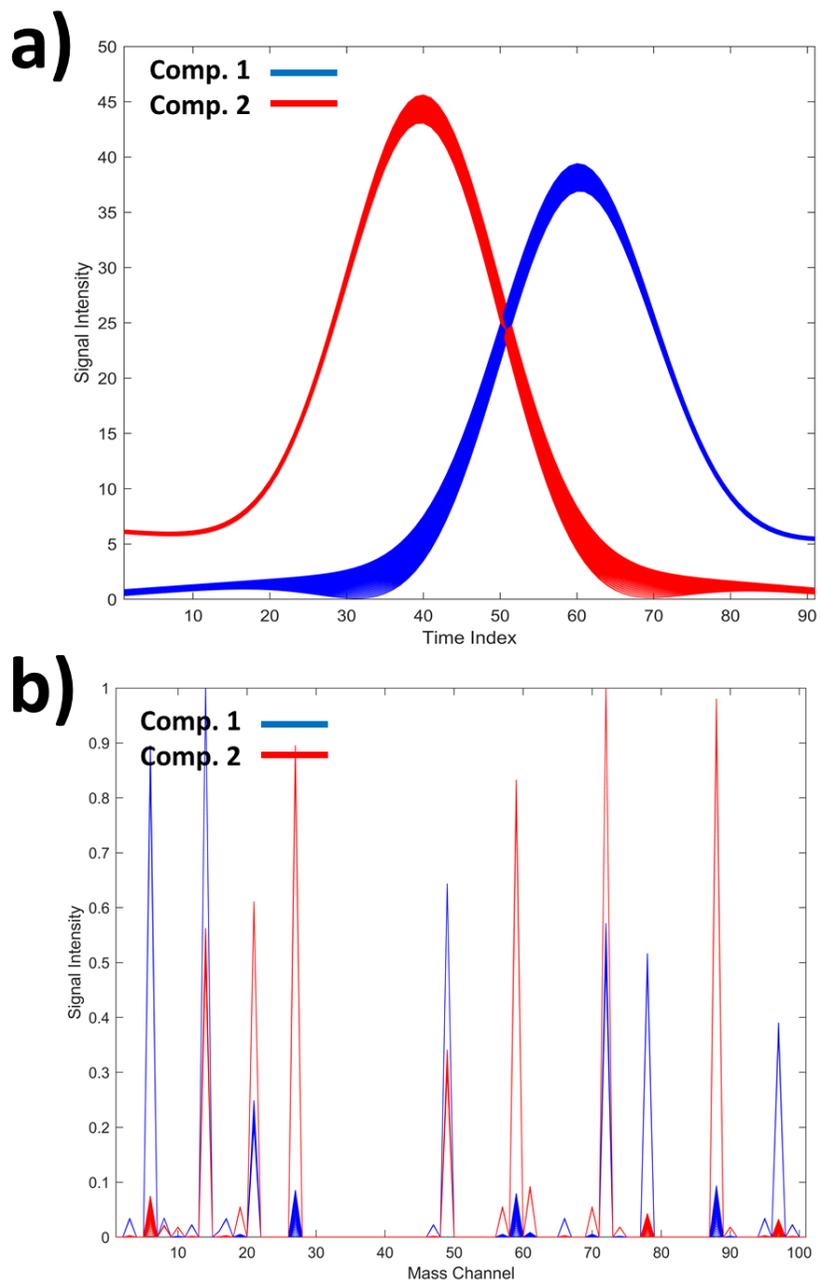

**Figure A4.** (a) The collection of concentration profiles and (b) the mass spectra which fits the data and non-negativity constraint for the simulated two component LC/GC-MS data in Appendix A.

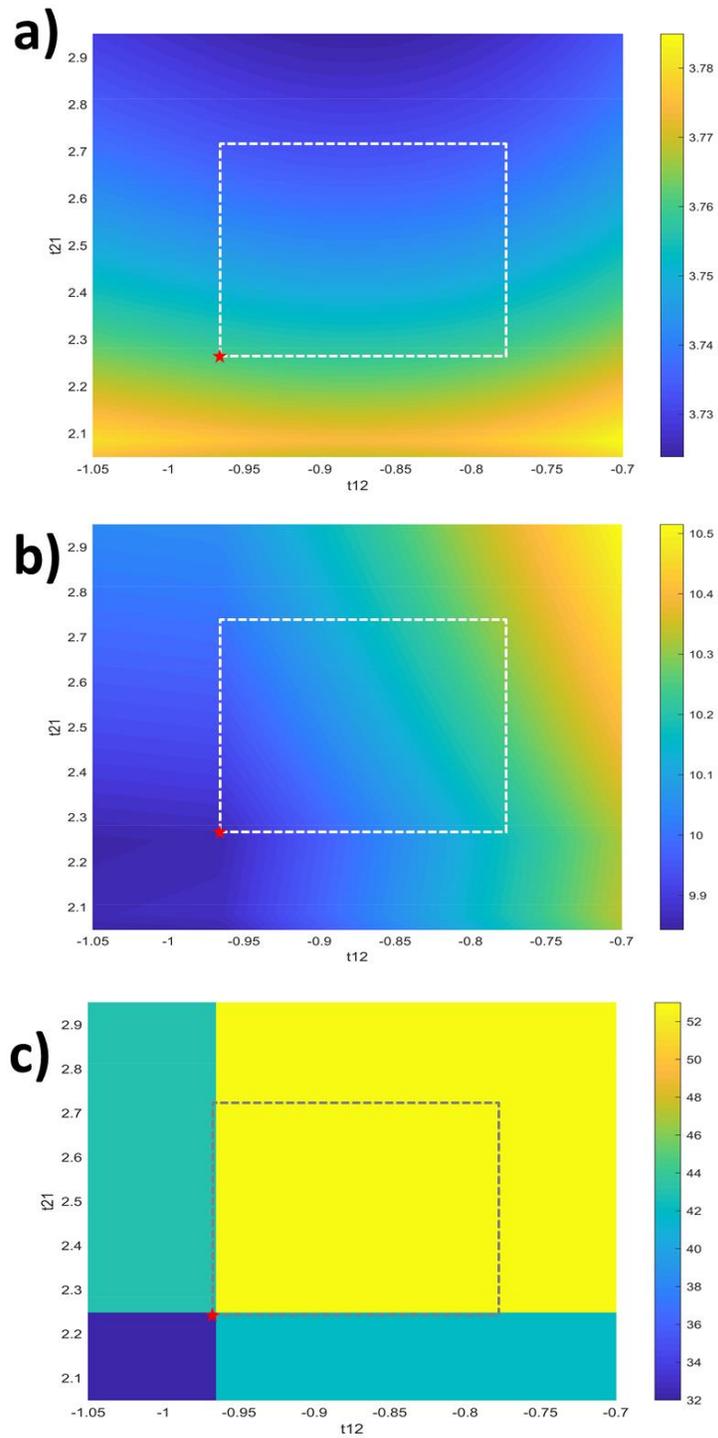

**Figure A5**. The contour plots for (a) $\sum L_{2^-}$, (b) $\sum L_{1^-}$ and (c) $\sum L_{0^-}$ norms in $t_{12}$-$t_{11}$ space for AFSI for the simulated two component LC/GC-MS data in Appendix A.

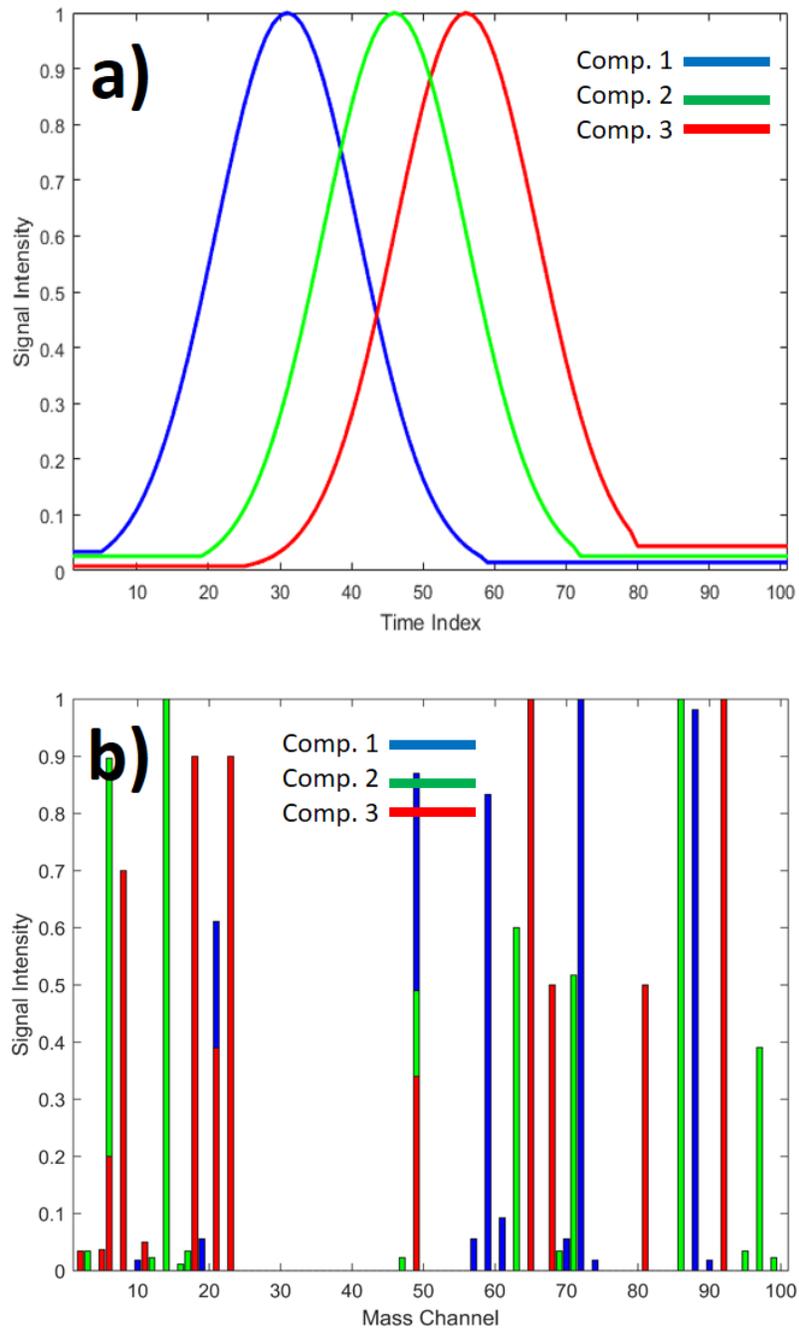

**Figure B1**. (a) The concentration profiles and (b) mass spectra for the simulated three component LC/GC-MS data in Appendix B.

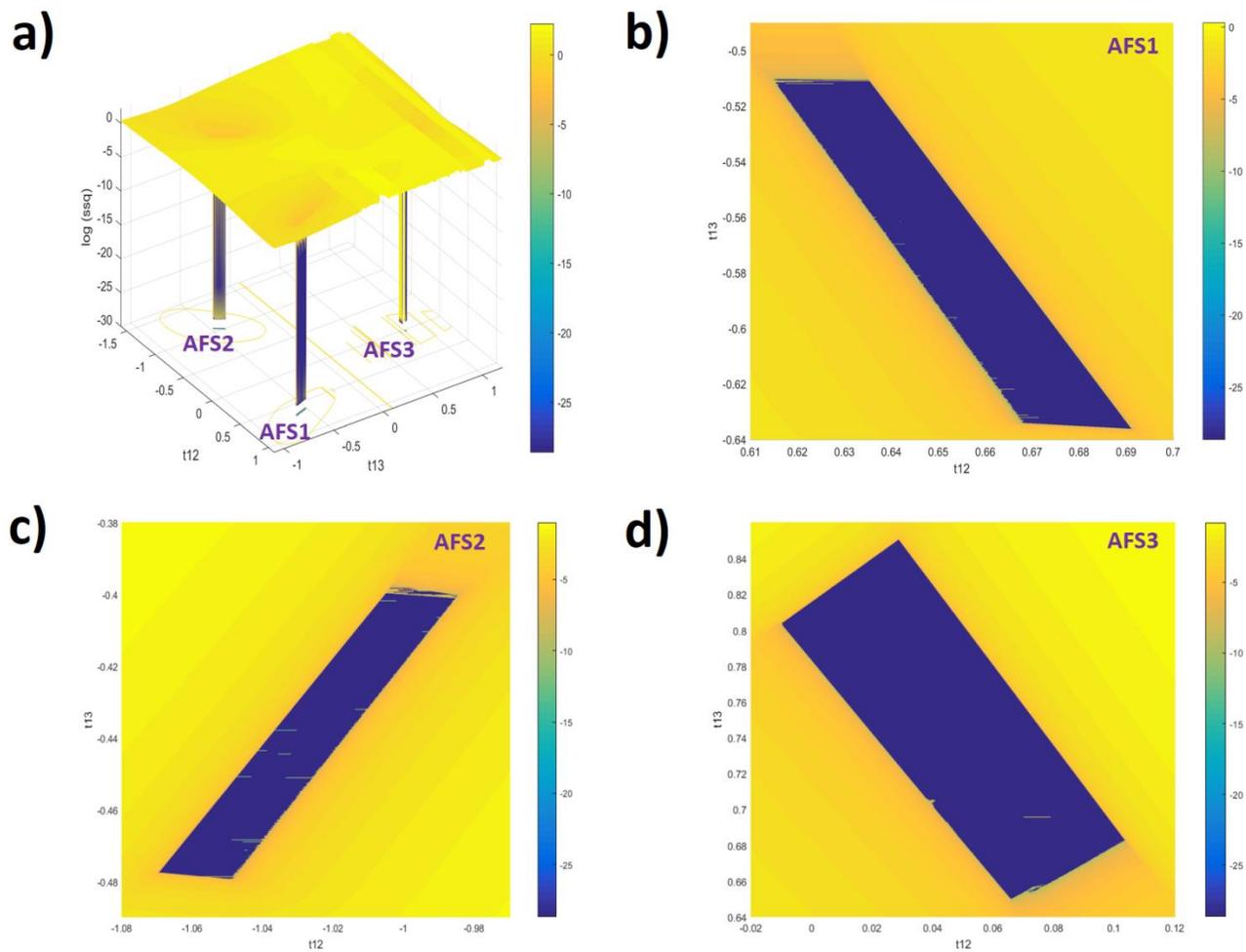

**Figure B2**. (a) The error surfaces obtained using grid search strategy and fminsearch approach for implementation of "non-negativity" constraint for the simulated three component LC/GC-MS data in Appendix B. (b) Contour plot of AFSI, (c) Contour plot of AFSII and (d) Contour plot of AFSIII

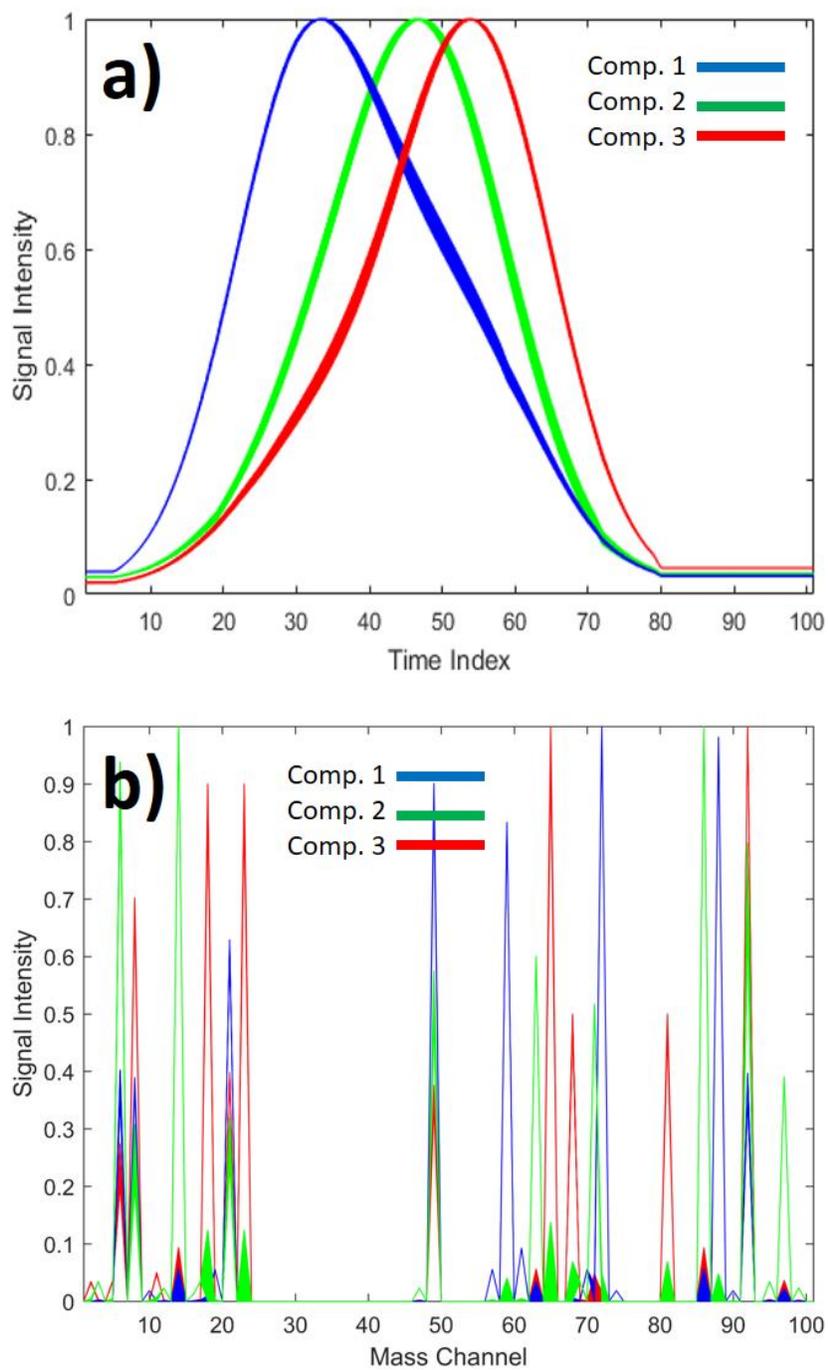

**Figure B3**. (a) The collection of concentration profiles and (b) the mass spectra which fits the data and non-negativity constraint for the simulated three component LC/GC-MS data in Appendix B.

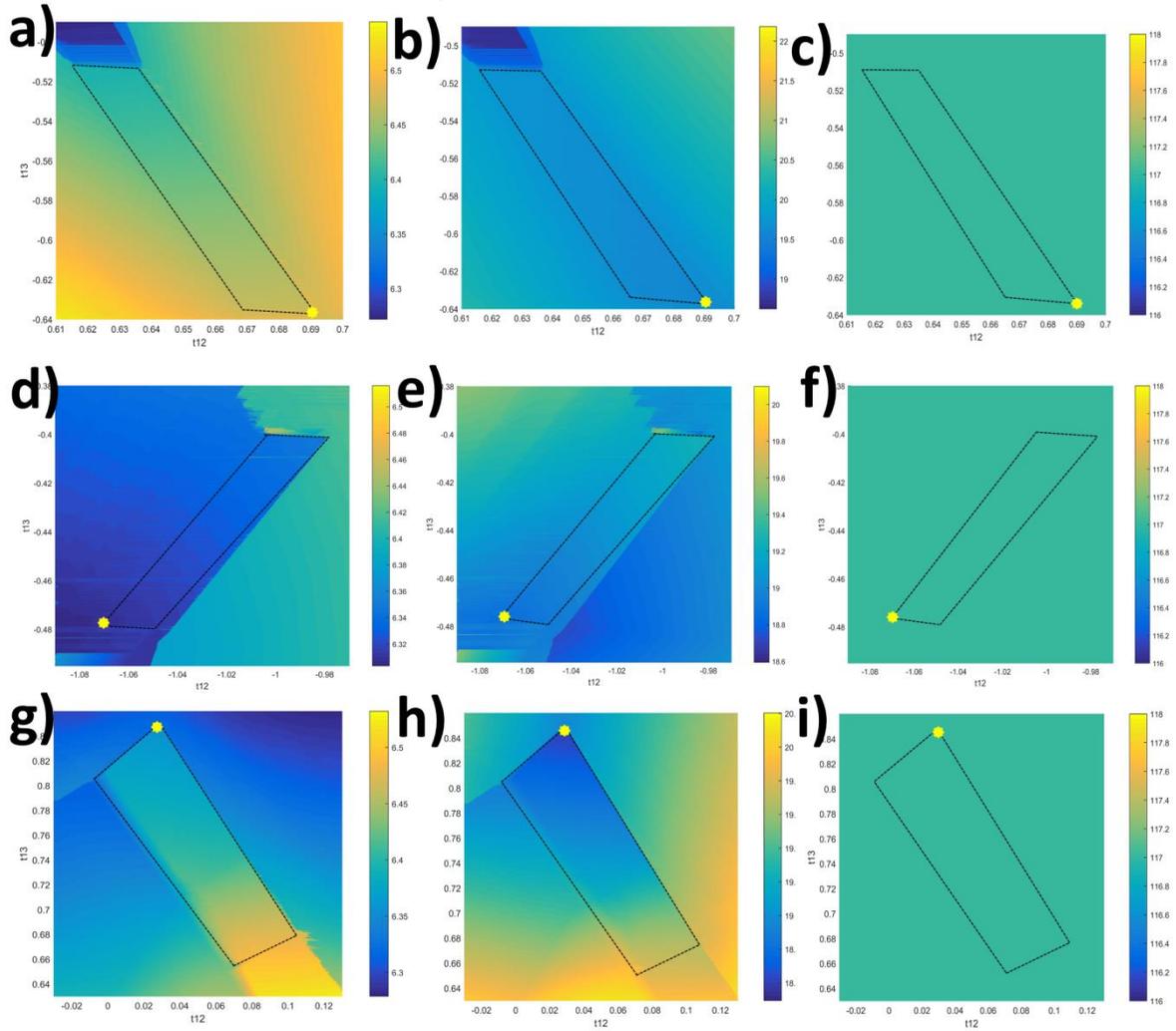

**Figure B4**. Contour plots for (a) $\sum L_2$-, (b) $\sum L_1$- and (c) $\sum L_0$- norms in $t_{12}$-$t_{13}$ space for AFSI, (d) $\sum L_2$-, (e) $\sum L_1$- and (f) $\sum L_0$- norms in $t_{12}$-$t_{13}$ space for AFSII and (g) $\sum L_2$-, (h) $\sum L_1$- and (i) $\sum L_0$- norms in $t_{12}$-$t_{13}$ space for AFSIII for the data simulated in Appendix B.

# Supplementary Material Section

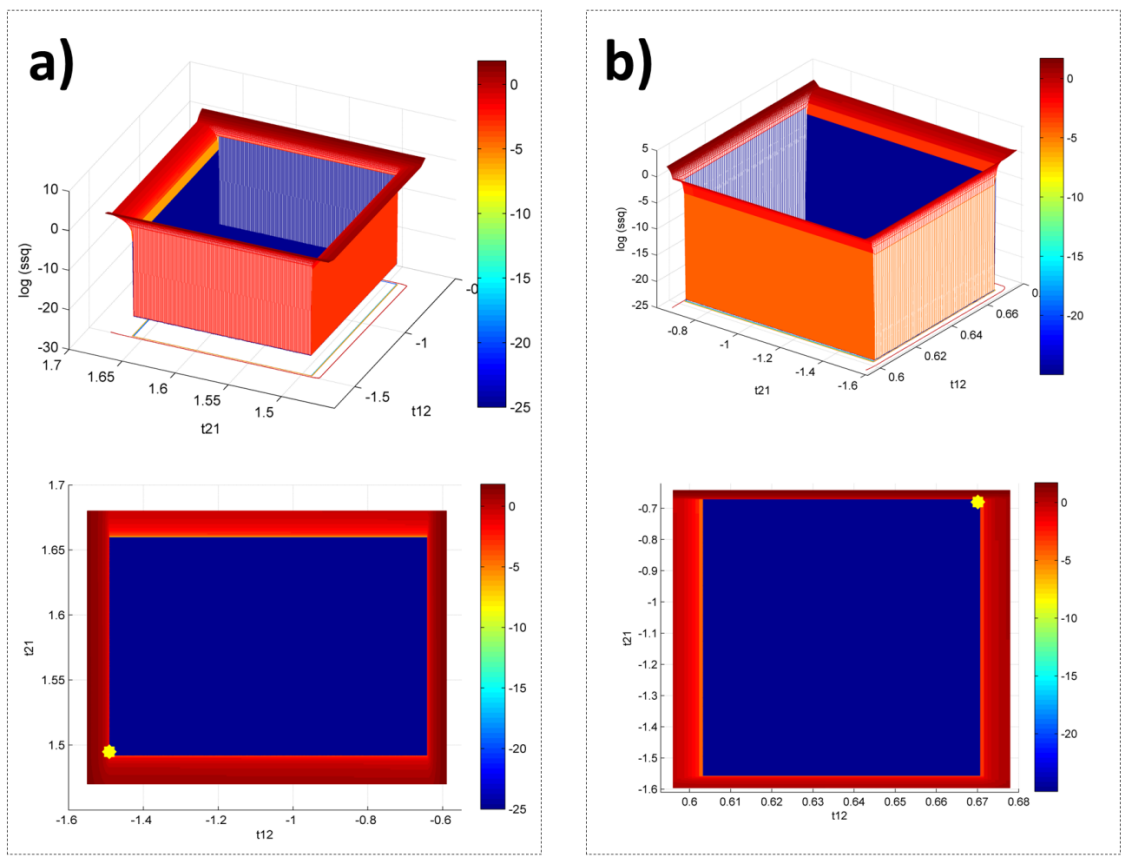

**Figure S1**. The plot of the log (ssq) against $t_{12}$ and $t_{21}$ together with their contour plots for the simulated two component system: (a) AFSI and (b) AFS II. Yellow star points reveal the true solutions.

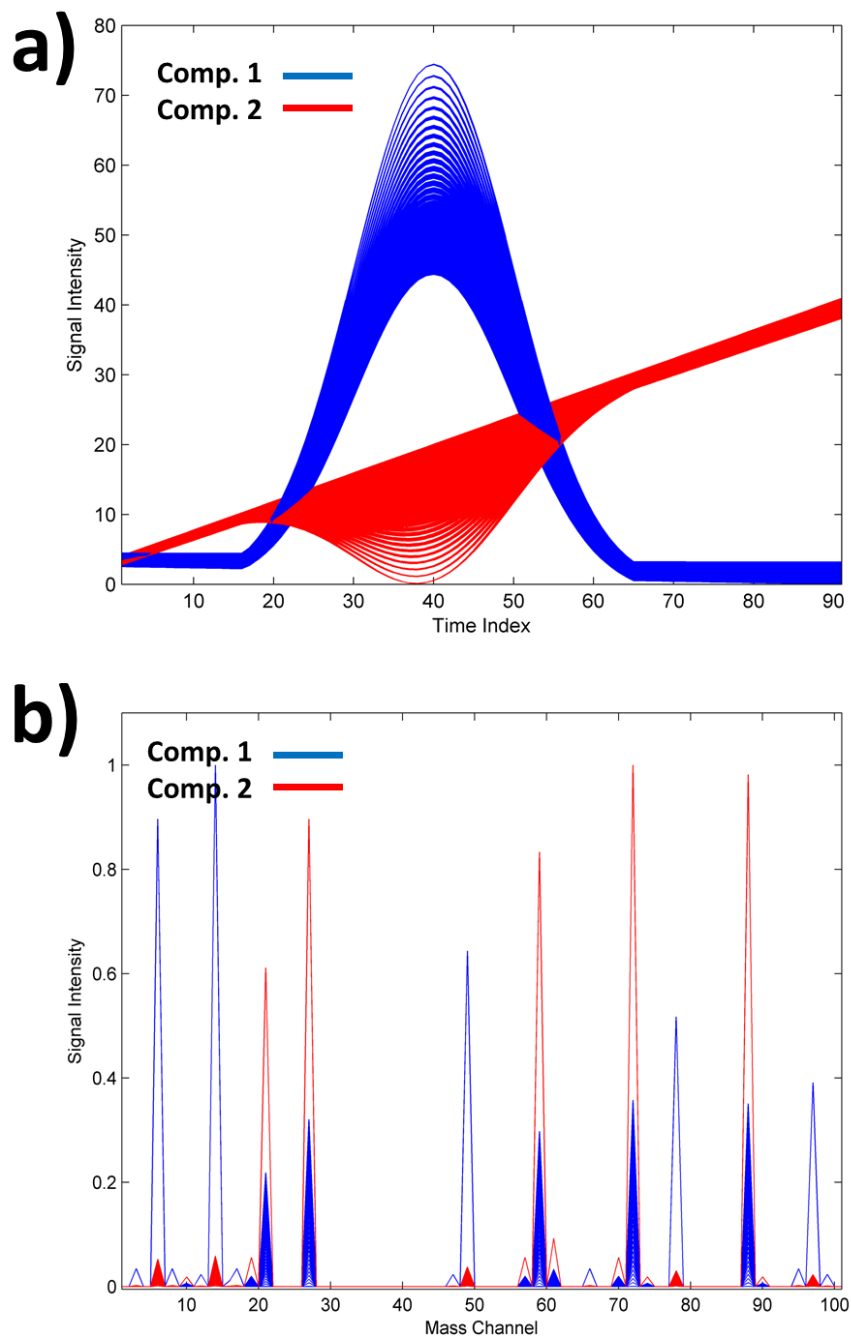

**Figure S2**. (a) The collection of concentration profiles and (b) the mass spectra which fits the data and non-negativity constraint for the simulated two component LC/GC-MS data.

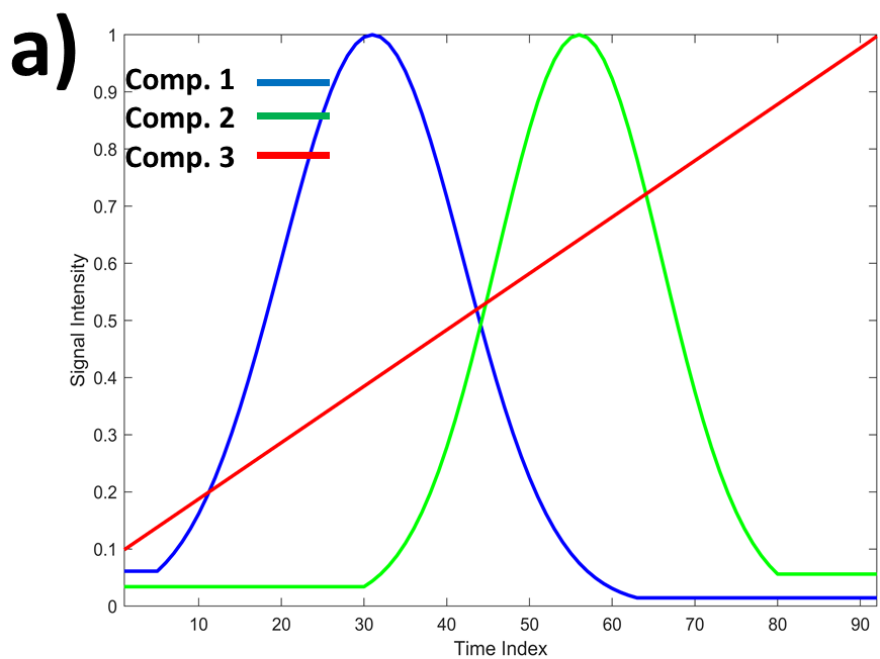
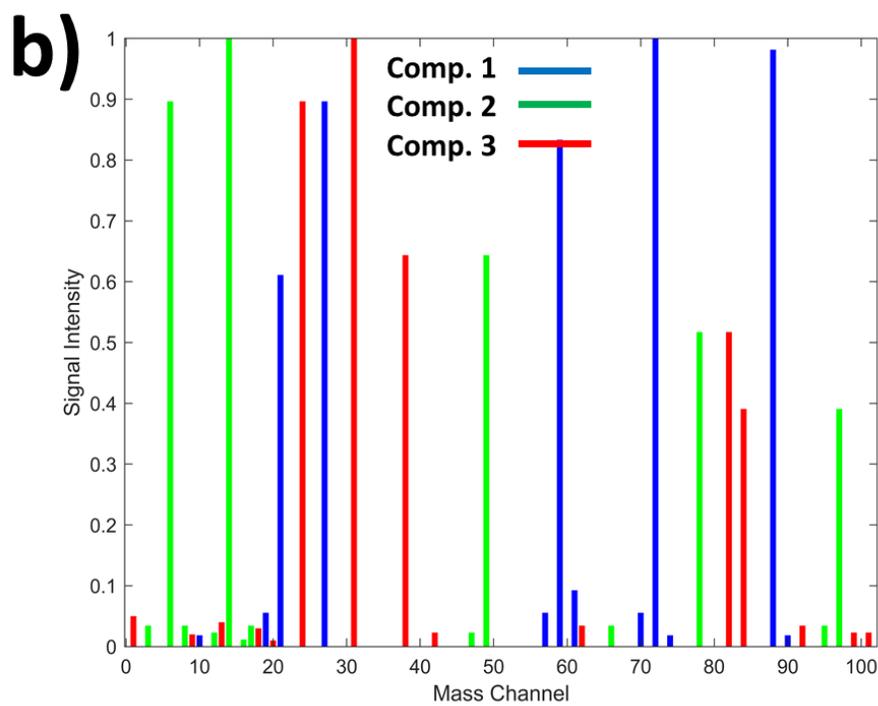

**Figure S3**. (a) The concentration profiles and (b) mass spectra for the simulated three component LC/GC-MS data.

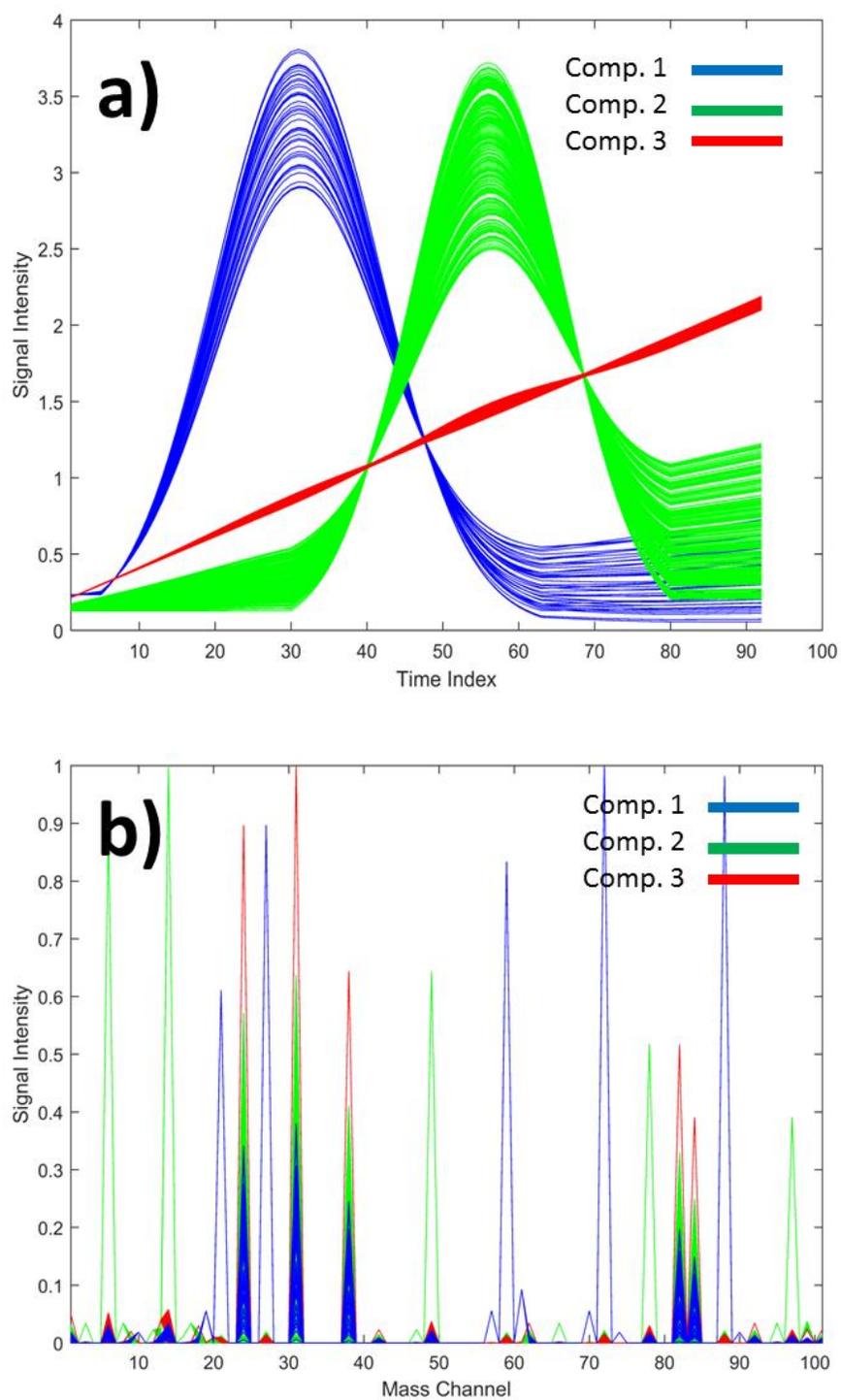

**Figure S4**. (a) The collection of concentration profiles and (b) the mass spectra which fits the data and non-negativity constraint for the simulated three component LC/GC-MS data

# High Resolution Images

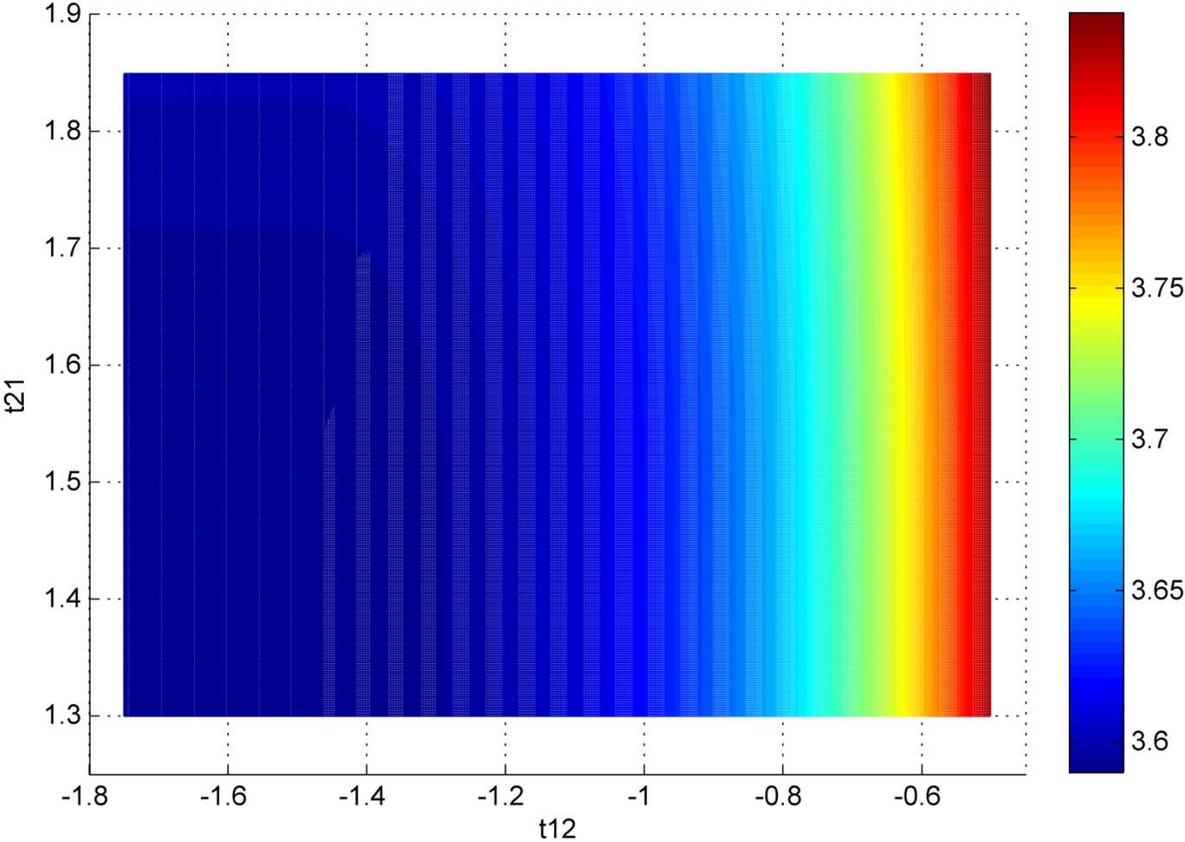

High resolution version of Figure 4a.

# High Resolution Images

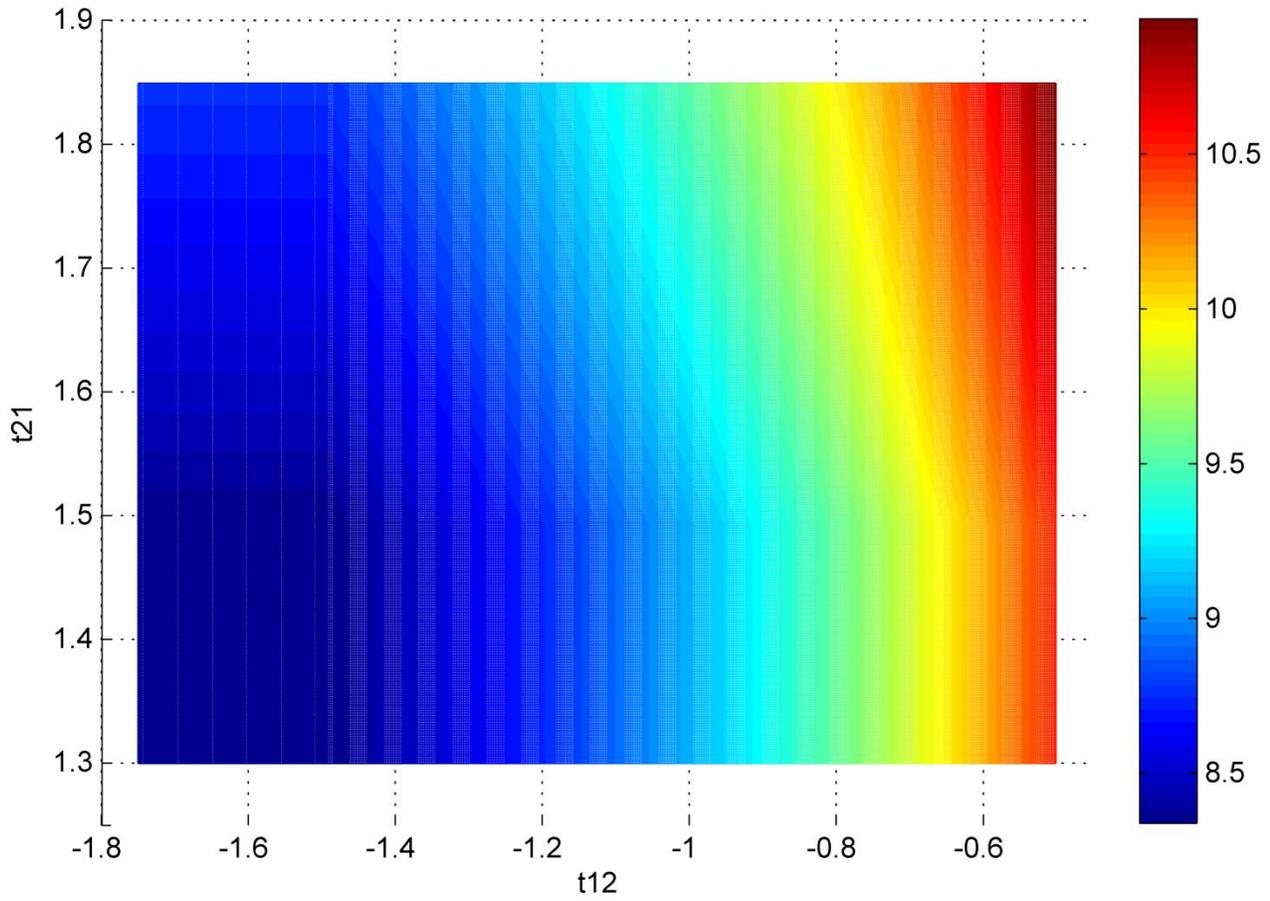

High resolution version of Figure 4b.

# High Resolution Images

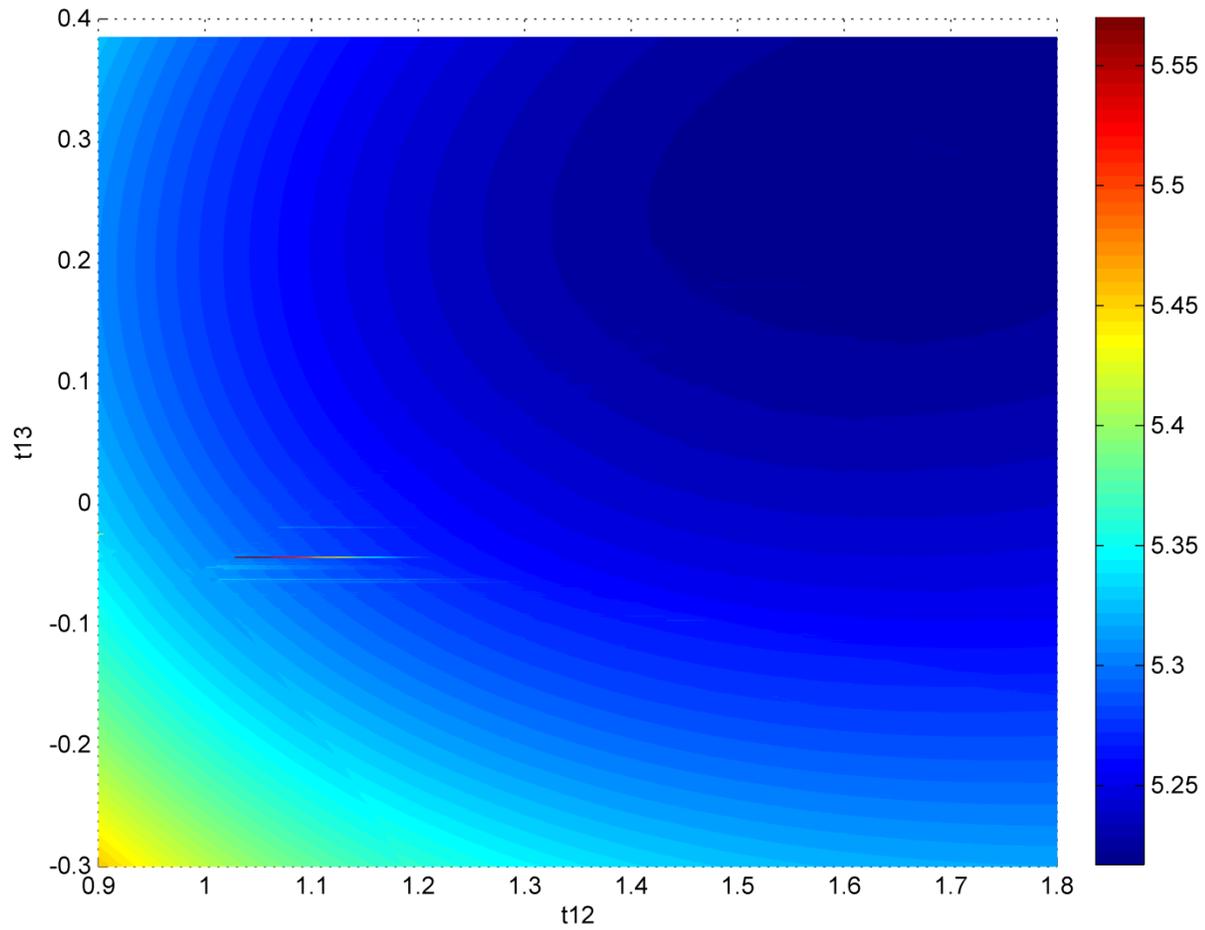

High resolution version of Figure 6a.

**High Resolution Images**

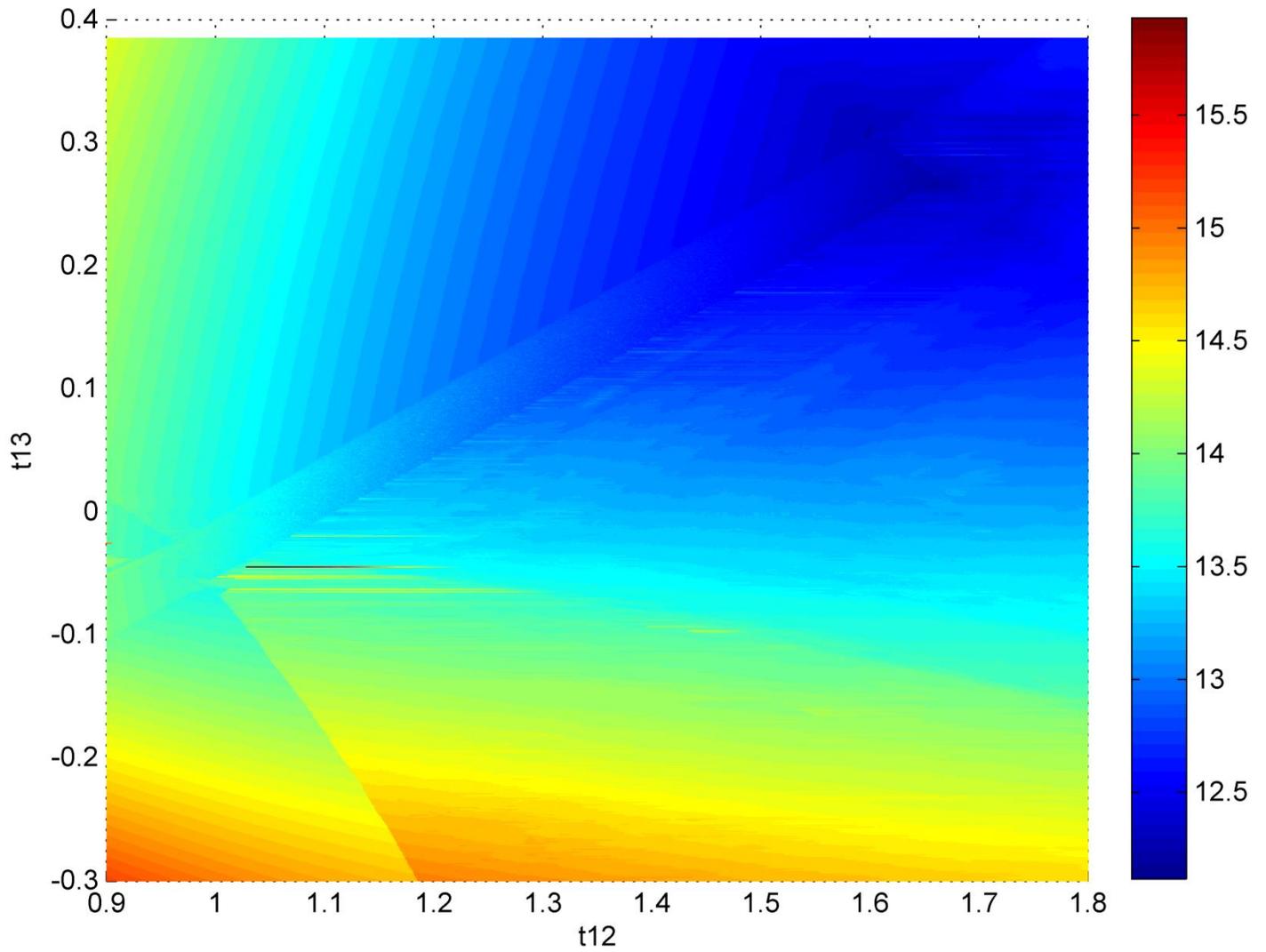

High resolution version of Figure 6b.

**High Resolution Images**

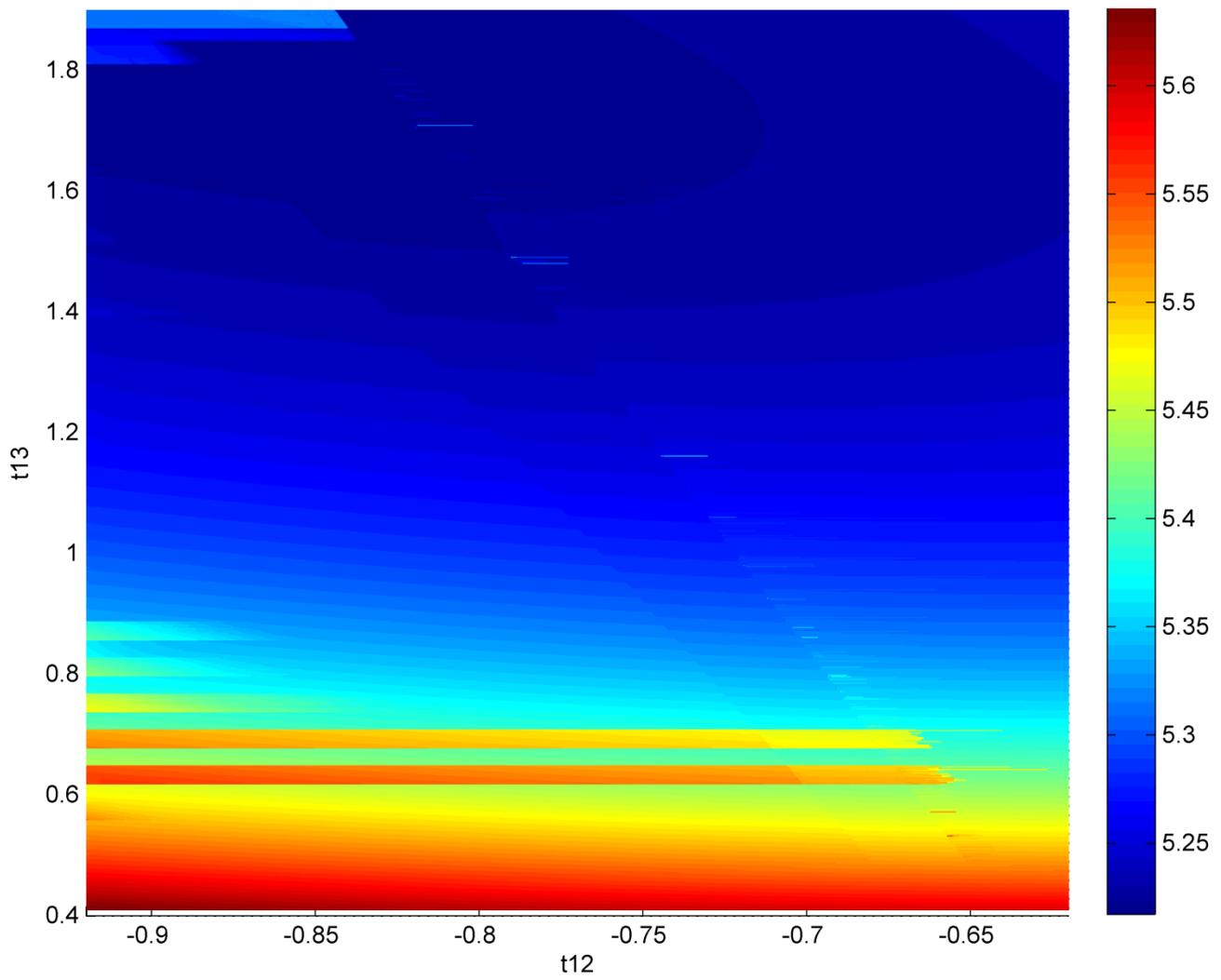

High resolution version of Figure 6d.

**High Resolution Images**

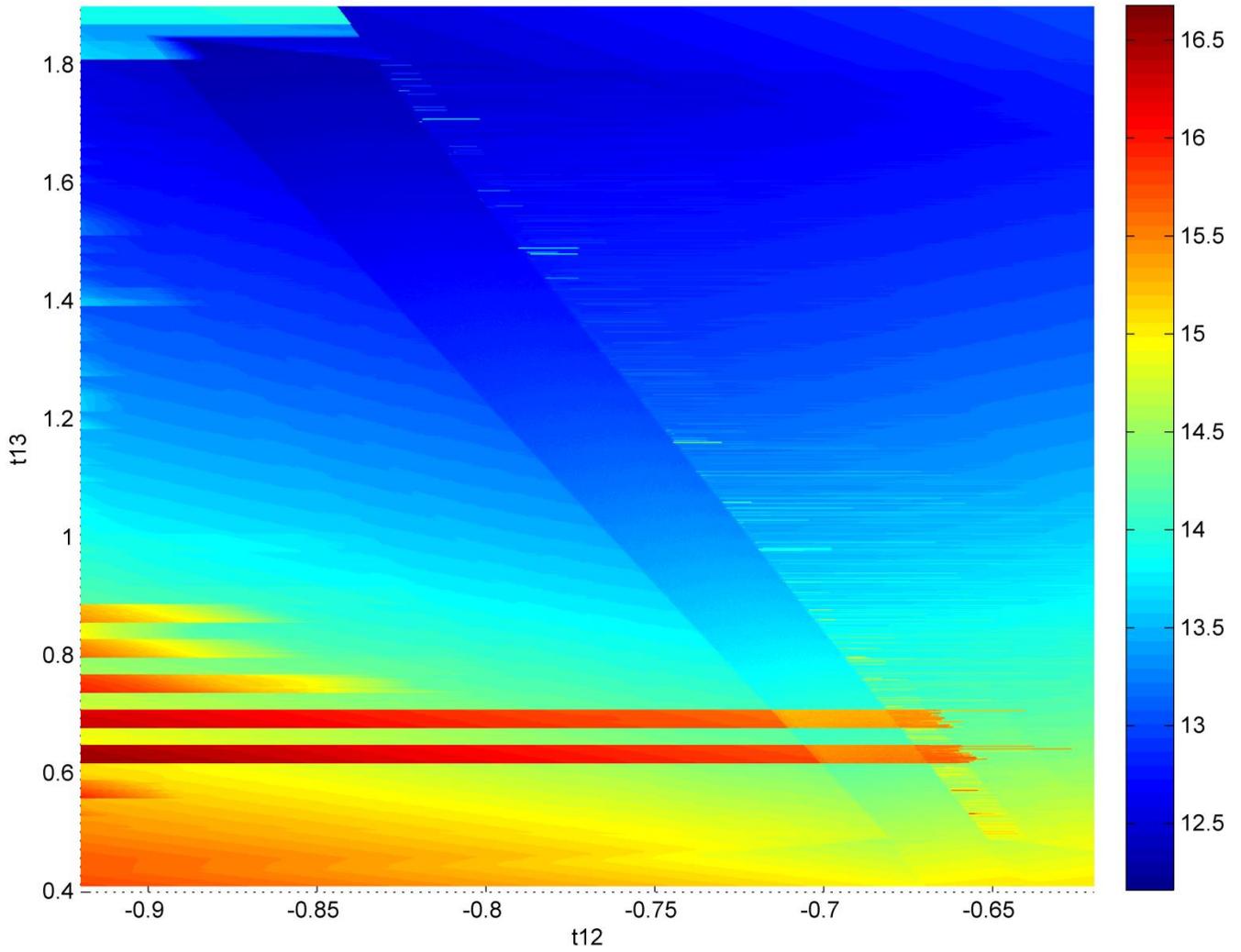

High resolution version of Figure 6e.

## High Resolution Images

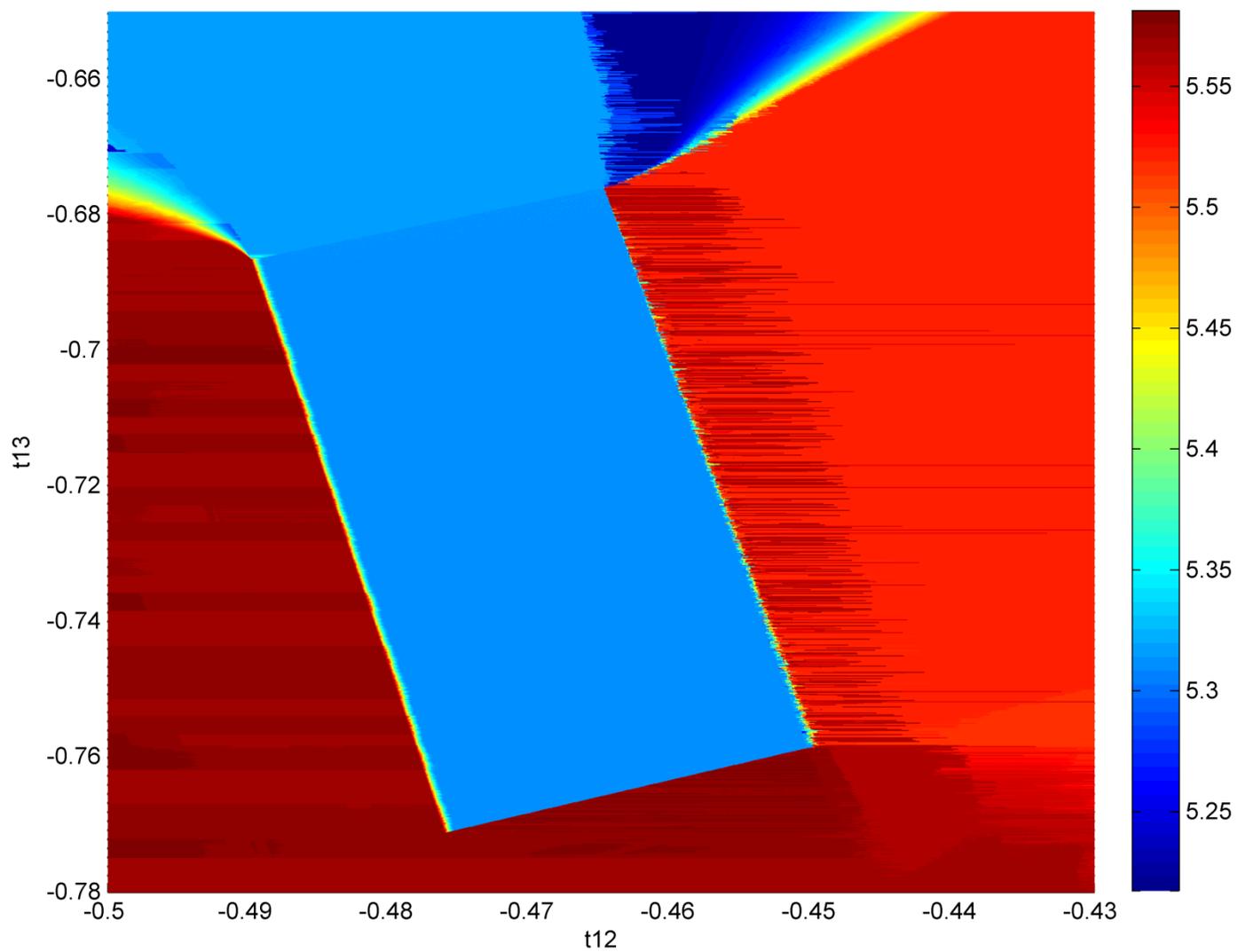

High resolution version of Figure 6g.

**High Resolution Images**

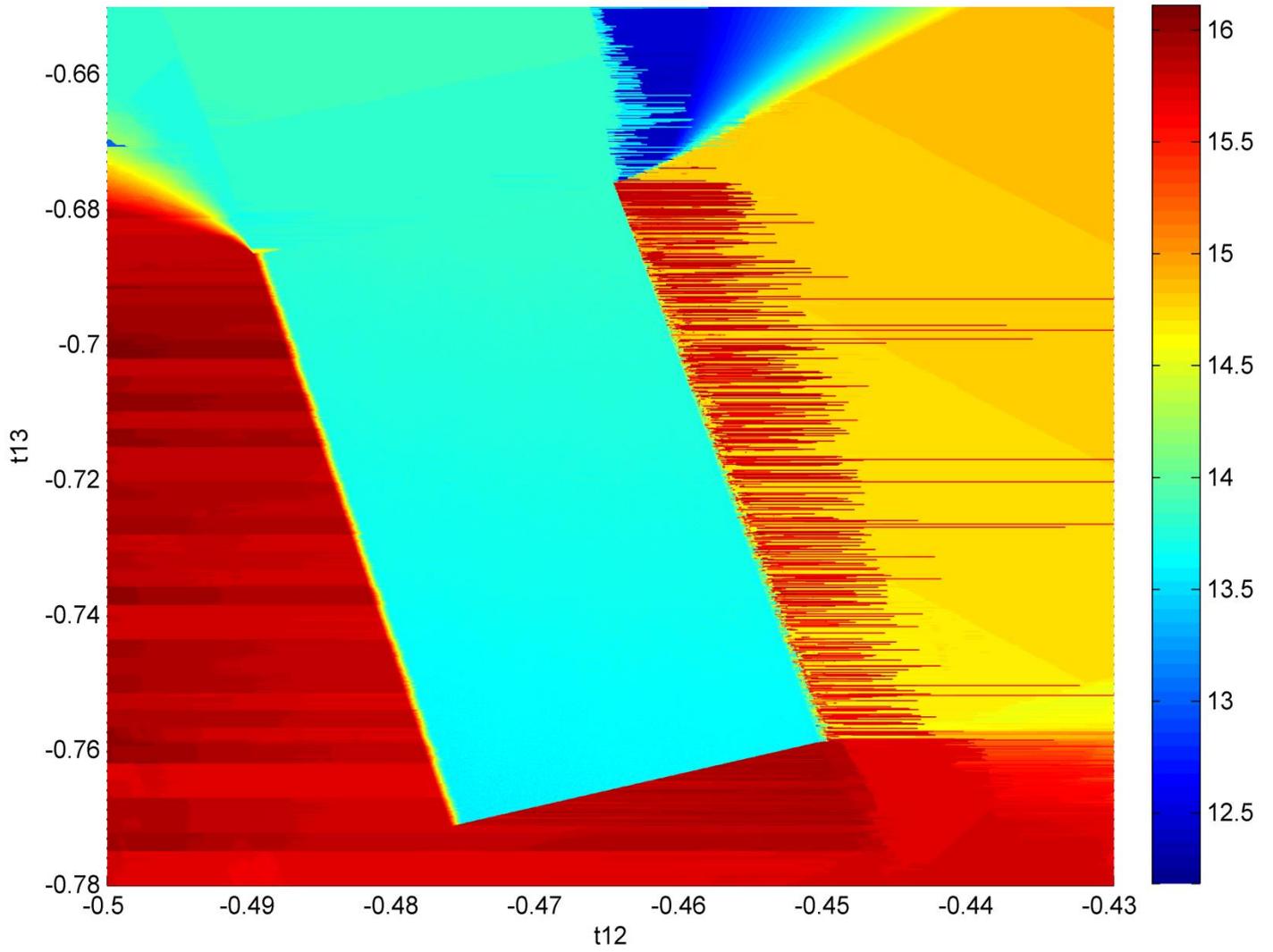

High resolution version of Figure 6h.